\documentclass[sigconf, nonacm]{acmart}

\newcommand\vldbdoi{XX.XX/XXX.XX}

\newcommand\vldbavailabilityurl{https://github.com/Bigscity-VecCity/VecCity}
\newcommand\vldbpagestyle{plain}

\usepackage[utf8]{inputenc} 
\usepackage[T1]{fontenc}    

\usepackage{url}            
\usepackage{booktabs}       
\usepackage{amsfonts}       
\usepackage{nicefrac}       
\usepackage{microtype}      
\usepackage{xcolor}         
\usepackage{setspace}
\usepackage{makecell}
\usepackage{svg}
\usepackage{colortbl}

\usepackage{array}  
\usepackage{lipsum}           

\usepackage{xspace}
\usepackage{bm}
\usepackage{amsmath}
\usepackage{multirow}
\usepackage{subfigure}
\usepackage{wrapfig}
\usepackage{graphicx}
\usepackage{amsthm}
\usepackage{booktabs}
\usepackage{pdflscape}

\usepackage[noend]{algpseudocode}
\usepackage{float}
\usepackage{hyperref}
\usepackage{algorithm}
\usepackage{algorithmicx}
\usepackage{enumitem}
\usepackage{pifont}
\usepackage[most]{tcolorbox}
\usepackage{placeins}

\usepackage{titletoc}

\usepackage[resetlabels]{multibib}
\newcites{New}{REFERENCES}

\newcommand{\paratitle}[1]{\vspace{1.1ex}\noindent\textbf{#1}}

\newcommand{\fig}{Fig.\xspace}

\theoremstyle{definition}
\newtheorem{definition}{Definition}

\newcommand{\ie}{\emph{i.e., }}
\newcommand{\name}{\emph{VecCity}\xspace}
\newcommand{\fullname}{the \emph{VecCity} library\xspace}

\newcommand{\eg}{\emph{e.g., }}

\newcommand{\etc}{\emph{etc}}

\newcommand{\rl}{MapRL\xspace}
\newcommand{\tax}{method-based taxonomy\xspace}

\newcommand{\git}{\hyperref[*]{\url{https://github.com/Bigscity-VecCity/VecCity}}}

\newcommand{\updateo}[1]{{\color{black} #1}}
\newcommand{\updatetw}[1]{{\color{black} #1}}
\newcommand{\updatetr}[1]{{\color{black} #1}}
\newcommand{\updatem}[1]{{\color{black} #1}}

\newcommand{\defaultstyle}{
    \renewcommand{\thesection}{\arabic{section}}
}

\begin{document}

\newpage

\defaultstyle 
\title{\name: A Taxonomy-guided Library for Map Entity Representation Learning [Experiment, Analysis \& Benchmark]}

\pagenumbering{arabic} 
\setcounter{page}{0} 
\setcounter{section}{0}

\author{Wentao Zhang$^1$, Jingyuan Wang$^1$, Yifan Yang$^1$, Leong Hou U$^2$}
\affiliation{%
  \institution{1. School of Computer Science and Engineering, Beihang University, Beijing, China \\
  2. Department of Computer and Information Science, University of Macau, Macau SAR, China}
}

\begin{abstract}
Electronic maps consist of diverse entities, such as points of interest (POIs), road segments, and land parcels, playing a vital role in applications like ITS and LBS. Map entity representation learning (\rl) generates versatile and reusable data representations, providing essential tools for efficiently managing and utilizing map entity data. Despite the progress in \rl, two key challenges constrain further development. First, existing research is fragmented, with models classified by the type of map entity, limiting the reusability of techniques across different tasks. Second, the lack of unified benchmarks makes systematic evaluation and comparison of models difficult.
To address these challenges, we propose a novel taxonomy for MapRL that organizes models based on functional modules—such as encoders, pre-training tasks, and downstream tasks—rather than by entity type. Building on this taxonomy, we present a taxonomy-driven library, \name, which offers easy-to-use interfaces for encoding, pre-training, fine-tuning, and evaluation. The library integrates datasets from nine cities and reproduces 21 mainstream MapRL models, establishing the first standardized benchmarks for the field. VecCity also allows users to modify and extend models through modular components, facilitating seamless experimentation. Our comprehensive experiments cover multiple types of map entities and evaluate 21 \name pre-built models across various downstream tasks. Experimental results demonstrate the effectiveness of \name in streamlining model development and provide insights into the impact of various components on performance. By promoting modular design and reusability, \name offers a unified framework to advance research and innovation in \rl. The code is available at \git.
\end{abstract}

\maketitle

 \pagestyle{\vldbpagestyle}


\setcounter{figure}{0}
\setcounter{table}{0}

\section{Introduction}


In the era of mobile internet, {\em electronic maps} have become a foundational platform for a wide range of applications, including intelligent transportation systems and location-based services. An electronic map consists of {\em map entities}, such as points of interest (POIs), road segments, and land parcels. These entities encapsulate complex geospatial data, spatial relationships, and geometric topological structures, presenting significant challenges for data representation. The effective representation of map entities has become a critical and enduring research focus within spatiotemporal data analysis and geographic information systems (GIS).

\updatem{Traditionally, map entities have been stored as structured records in relational databases or geographic file formats, such as PostGIS~\cite{postgis}, Shapefiles~\cite{shapefile}, and GeoJSON~\cite{geojson}. With this representation approach, services relying on map entities are required to develop task-specific models or algorithms to process the entities’ attributes and extract geographic relationships within or between them. However, these task-specific models lack generalizability, limiting their reusability across different applications and scenarios.}


\updatem{In recent years, pre-trained representation learning has emerged as a powerful approach for generating versatile, task-independent data representations. It has achieved remarkable success across various domains, including natural language processing (NLP)~\cite{gpt,elmo,glove} and computer vision (CV)~\cite{simclr,moco,vit}. The field of spatiotemporal data analysis has also embraced pre-trained representation learning to construct generalized representations of map entities ~\cite{location_encoding,chen2024self,ufm,opencity,uukg}, leading to the rise of an emerging research area known as Map Entity Representation Learning (\rl). However, despite rapid advancements, two major structural challenges require further exploration to unlock the potential of \rl.}

{\em Challenge 1: Fragmented Research Fields.} Electronic maps consist of various types of map entities, such as POIs (points), road segments (polylines), and land parcels (polygons). Existing research often treats the modeling of these entities as separate areas, limiting the reusability of techniques, methods, and modules across fields. However, real-world electronic map applications typically require the integrated use of all three entity types. This segregation not only creates obstacles for practical applications but also forms implicit academic exchange barriers between subfields, thereby constraining the progress of \rl research.

\updatem{{\em Challenge 2: Lack of Standardized Benchmark.} Despite numerous models having been proposed, their performance is often evaluated on different datasets and under varying experimental settings. Unlike the CV and NLP fields, this domain lacks standardized datasets and benchmarks. This absence hinders fair comparisons between models, making it difficult to derive generalizable design principles.}

\updatetr{
To address the first challenge, we propose a novel taxonomy for \rl models, namely \tax, which organizes the essential components of an \rl model into four key elements: {\em Map Data}, {\em Encoder Models}, {\em Pre-training Tasks}, and {\em Downstream Tasks}. Unlike traditional taxonomies that categorize models by map entity types, \tax takes a functional approach, grouping encoder models into three paradigms—{\em Token-based}, {\em Graph-based}, and {\em Sequence-based}—each supporting specialized pre-training tasks. Only downstream tasks depend on the nature of the map entity. A key insight of our taxonomy is that the two core elements—Encoder Models and Pre-training Tasks—are not tightly coupled with specific map entity types. This decoupling of core components from entity-specific constraints fosters a more unified and transferable framework for MapRL, facilitating cross-domain integration and knowledge sharing.
}

\updatem{To tackle the second challenge, we introduce \fullname, an \rl model development toolkit guided by our taxonomy. The library organizes map entities and auxiliary data into three atomic file types. Using these atomic files, we format data from nine cities into a unified structure. VecCity further standardizes model implementation with four interface functions covering encoders, pre-training tasks, and downstream tasks. Using these interfaces, we reproduce 21 mainstream \rl models and conduct evaluations across city datasets, providing key insights for future research.}

The main contributions of this paper are as follows:

$\bullet$ {\em Novel Taxonomy.} We present a novel taxonomy of \rl models that goes beyond the traditional classification based on map entities, offering a unified classification system applicable to \rl models across various types of map entities.

$\bullet$ {\em Easy-to-use Toolkit Library.} Guided by our taxonomy, we propose a toolkit, \name, which unifies the development process of \rl models. To the best of our knowledge, \name is the first unified model toolkit specifically designed for \rl.

$\bullet$ {\em Standard Benchmark.} \name implements 21 mainstream \rl models and evaluates them on datasets from nine cities, establishing the first unified benchmark for \rl models and offering valuable insights to guide future research.



\vspace{-0.1cm}
\section{Taxonomy of \rl} 
\label{sec:taxonomy}

\vspace{-0.1cm}

In this section, we present the proposed \tax. Figure~\ref{fig:generic} (a) outlines the general framework for \rl. The training process of a typical \rl model consists of two key stages: {\em pre-training} and {\em fine-tuning}. During pre-training, the \rl model takes map data as input, using an encoder model to transform them into representation vectors. The encoder's parameters are optimized through pre-training tasks. In the fine-tuning stage, downstream tasks are employed to further refine the encoder. This pipeline defines four core components of an \rl model: {\em Map Data}, {\em Encoder Models}, {\em Pre-training Tasks}, and {\em Downstream Tasks}.

\subsection{Map Data}~\label{sec:taxonomy_data}
\vspace{-0.2cm}

The map data consists of map entities and auxiliary data.


\subsubsection{Map Entities}\;

\begin{figure}[t]
    \centering
    \includegraphics[width=\columnwidth]{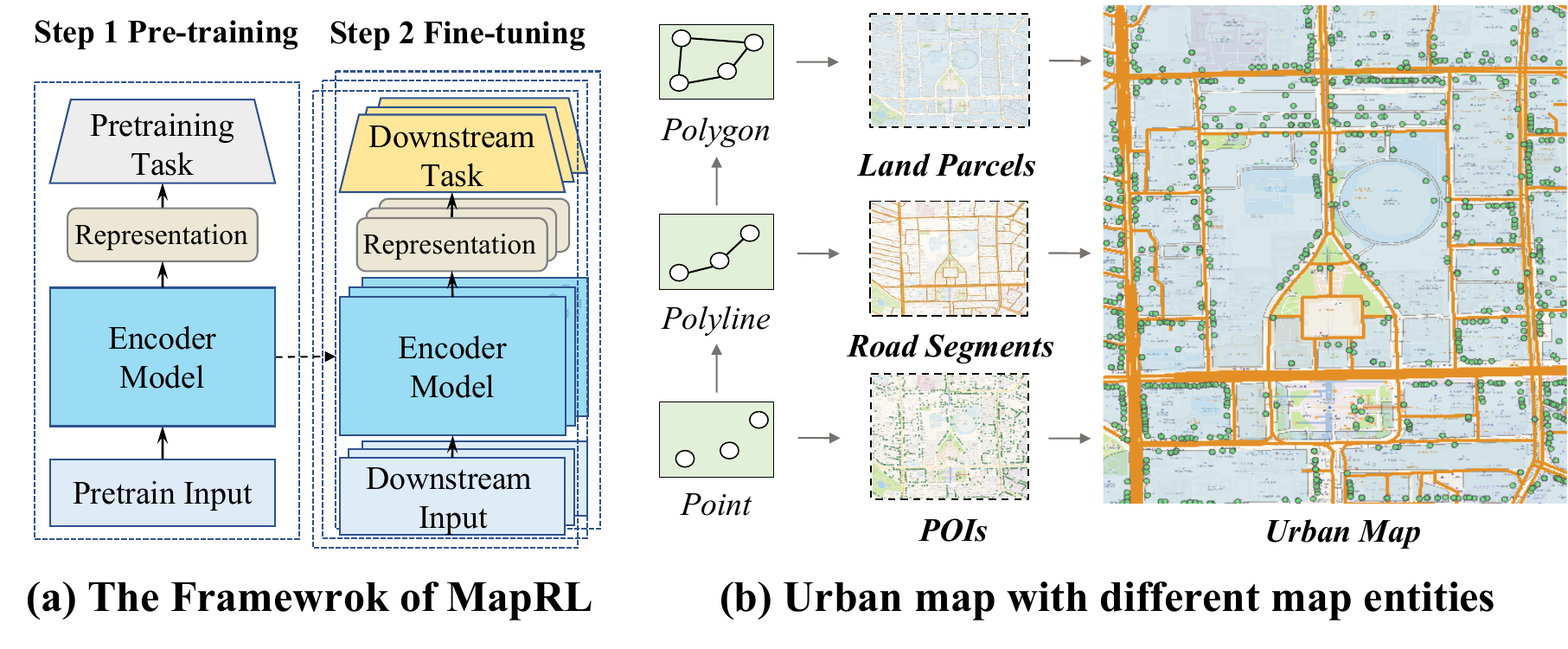}
    \vspace{-0.5cm}
    \caption{\updatem{Illustration of (a) the framework of a \rl model and (b) an urban map with different map entities.}}
    \label{fig:generic}
    \vspace{-0.4cm}
\end{figure}

As illustrated in Fig.~\ref{fig:generic} (b), an electronic map consists of three geometric elements: {\em Points}, {\em Polylines}, and {\em Polygons}, each representing a distinct type of map entity.\label{sec:test}

$\bullet$ {\em Points / POI.} A point element represents a {\em Point of Interest (POI)}, identified by a location coordinate, \ie $(latitude, longitude)$.

$\bullet$ {\em Polylines / Road Segment.} A polyline element represents a {\em Road Segment}, composed of a sequence of connected line segments, with each endpoint defined by location coordinates.

\updatem{$\bullet$ {\em Polygons / Land Parcel.} A polygon represents a {\em Land Parcel}, delineated by a closed sequence of connected line segments.}

For each map entity, a set of features describes its properties. For instance, the category feature of a POI differentiates the services it offers. Road segments are characterized by features like speed limits, lane counts, \etc. Based on these definitions, we formally define electronic maps and map entities as follows.


\begin{definition}[Electronic Maps and Map Entities]
An electronic map is defined as a set of map entities, $V = \{v_1, \ldots, v_i, \ldots, v_I\}$, where each entity $v_i$ consists of four components: {\em ID}, {\em type}, {\em geometric shape}, and {\em feature}. The {\em ID} serves as the unique identifier of $v_i$. The {\em type} specifies the entity's geometric form -- point, polyline, or polygon. The {\em geometric shape} provides the coordinates that define its structure, while the {\em feature} captures its various attributes.
\end{definition}


\subsubsection{Auxiliary Data}\;

\vspace{1mm}
Besides map entities, two types of auxiliary data are commonly used in \rl models: {\em Trajectories} and {\em Relation Networks}.

\begin{definition}[Trajectories]
Trajectories refer to the mobility records of humans or vehicles. A trajectory is defined as a sequence of locations with corresponding timestamps, \ie  $tr = (<l_{tr_1},t_{tr_1}>, \ldots,  <l_{tr_k},t_{tr_k}>, \ldots,  <l_{tr_K},t_{tr_K}>)$, where $l_{tr_k}$ and $t_{tr_k}$ are the location and timestamp for the $k$-th trajectory sample.
\end{definition}

Trajectory data can be divided into two classes based on the type of location samples $l_{tr_k}$. $i$) {\em Check-in Trajectories}: These use POIs as samples and are collected through social networks or LBS apps. They log a user's location only upon check-in, resulting in a low sampling rate, often requiring days to capture a single point. \rl models leverage these trajectories to infer users' personal preferences. $ii$) {\em Coordinate Trajectories}: These rely on geographic coordinates as samples. Collected by ITS through GPS devices or mobile phones, they provide high-frequency sampling along with real-time speed and direction data. These trajectories are aligned with road segments using map matching algorithms~\cite{fmm} and facilitate route planning and dynamic state estimation for road segments.

\begin{definition}[Relation Network]~\label{def:Relation_Networks}
A relation network represents map entities as a directed graph $G = \{V, \bm{E}\}$, where $V = \{v_1, \ldots, v_I\}$ represents the set of vertices corresponding to map entities. The adjacency matrix $\bm{E} \in \mathbb{R}^{I \times I}$ defines the relationships between entities, with each element $e_{ij}$ indicating the relationship from $v_i$ to $v_j$.
\end{definition}

Network relations can be categorized into two types: $i$) {\em Geographical Relations (GR)}: These leverage spatial relationships between map entities to construct the adjacency matrix. Examples include connections between road segments in a road network~\cite{HyperRoad} and distances between POIs~\cite{Teaser,DLCL}. $ii$) {\em Social Relations (SR)}: These represent user behaviors, such as mobility trajectories, which are used to construct adjacency matrices. A typical example is {\em Origin-Destination (OD) flows}, widely used in \rl models~\cite{MVURE,HREP}, where $e_{ij}$ in the adjacency matrix represents the number of trajectories that start from $v_i$ and end at $v_j$.

\subsection{Encoder Models and Pre-training Tasks}~\label{sec:taxonomy_pretraining}

Encoder models and pre-training tasks are two core components of a pre-trained \rl model, responsible for converting input map data into representation vectors for map entities. Our taxonomy classifies encoder models into three types: {\em token-based}, {\em graph-based}, and {\em sequence-based}, designed to model features, relation networks, and trajectory sequences of map entities, respectively. 

\vspace{-4mm}
\subsubsection{Token-based Models}\;

\paratitle{Encoders.} The token-based encoder generates representation vectors directly from the features of map entities. For discrete features, it applies one-hot encoding, and for continuous features, the value range is divided into consecutive bins, each represented by a one-hot code. Given $K$ features for a map entity, let the one-hot code of the $k$-th feature be $\bm{f}_k \in \{0,1\}^{F_k}$ (where $F_k$ is the dimensionality). \updatem{The encoder maps each feature to a learnable embedding space using a matrix $\bm{R}_k \in \mathbb{R}^{F_k \times d}$, generating the entity representation as $\bm{r} = \big\|_{k=1}^K \bm{f}_k^{\top} \cdot \bm{R}_k \big.$, where $\|_{k=1}^K$ denotes the concatenation of embeddings from all features.}

\updatem{Existing token-based methods primarily innovate by refining how they extract meaningful features from map data, which can be classified into three key types: $i$) {\em Spatial Features}: These capture the locations and spatial structures of map entities. The most fundamental spatial feature is coordinate information. Hier~\cite{Hier1} enhances spatial representation by embedding hierarchical spatial grids into rasterized maps. $ii$) {\em Temporal Features}: The characteristics of a map entity can vary across different times of the day. For instance, a restaurant's activity level differs between working hours and lunchtime. To model such variations, many token-based encoders divide the day into segments (e.g., hourly or half-hourly) and assign each segment a unique embedding to capture temporal dynamics~\cite{TALE,Teaser,CACSR,JGRM,CWAP}. $iii$) {\em Semantic Features}: These reflect the intrinsic properties of map entities, such as the functional types of POIs or the road types of segments. Some land parcel \rl methods further enhance semantic features by counting the number of POI categories within a parcel~\cite{REMVC}.}

\paratitle{Pre-training Tasks.} For token-based encoders, the corresponding pre-training tasks include three types:

$\bullet$ {\em Token Relation Inference (TokRI).} This task aims to predict the relationship between two map entities based on their representation vectors. Specifically, given the representations $\bm{r}_i$ and $\bm{r}_j$ of two entities, the TokRI task predicts their relationship as $\hat{y}_{ij} = \phi(\bm{r}_i, \bm{r}_j)$, where $\phi(\cdot)$ is the prediction function, and $\hat{y}_{ij} \in [0,1]$ is the predicted probability of a relation existing. The encoder parameters are optimized during pre-training using cross-entropy loss.

$\bullet$ {\em Token Relation Contrastive Learning (TRCL).} This task ensures that entities with closer relationships are more likely to have similar representations. Given a representation sample $\bm{c}_i$, another sample $\bm{c}_j$ is treated as a positive sample if the corresponding entities $v_i$ and $v_j$ share a close relation; otherwise, $\bm{c}_j$ is treated as a negative sample. The encoder parameters are optimized using a contrastive learning loss, such as the InfoNCE loss~\cite{InfoNCE}.

$\bullet$ {\em Augmented Token Contrastive Learning (AToCL).} These tasks use data augmentation techniques to generate different views of a given entity's representation. A contrastive learning loss is then applied to maximize the similarity between views of the same entity while minimizing the similarity between different samples.\label{sec:t221}

\updateo{For TokRI and TRCL tasks, the primary innovation in existing models lies in how they construct relations. These methods can be categorized into three classes: $i$) {\em Distance-Based Relations}: This approach establishes relations based on spatial proximity. In TokRI tasks, RN2Vec~\cite{RN2Vec} defines a relationship between entities if their geographical distance falls below a predefined threshold. In TRCL tasks, Teaser~\cite{Teaser} selects the $K$ nearest entities as positive samples for contrastive learning. $ii$) {\em Semantic-Based Relations}: Senmant~\cite{semantic}, Place2V~\cite{Place2Vec}, and MT-POI~\cite{MT-POI} select entities that have the same or similar semantic features as positive samples for contrastive learning. $iii$) {\em Auxiliary Data-Based Relations}: This method leverages additional data, such as trajectories, to define relations. In TokRI tasks, P2Vec~\cite{POI2VEC} and Tale~\cite{TALE} treat a trajectory as a sentence and the entities within it as words, using the Continuous Bag of Words (CBOW) model~\cite{Word2Vec} to learn representations for ``words'' (map entities). In TRCL tasks, CatEM~\cite{CatEM}, R2Vec~\cite{r2vec}, and Teaser~\cite{Teaser} employ the SkipGram model~\cite{Word2Vec} with negative sampling to capture contextual relations along trajectories. When trajectory data is unavailable, ``virtual trajectories'' can be generated via random walks on relation networks~\cite{HDGE}. Beyond trajectories, ZEMob~\cite{ZEMob} uses OD flows between entities to identify relationships.}

For AToCL tasks, the innovation primarily lies in the methods used to generate augmented samples. For instance, ReMVC~\cite{REMVC} and ReDCL~\cite{RegionDCL} enhance contrastive learning by replacing or dropping specific features of the anchor entity to create augmented variants.

\subsubsection{Graph-based Models}\;


\paratitle{Encoders.} Graph-based encoders leverage auxiliary data in the form of relation networks (see Definition~\ref{def:Relation_Networks}) to capture relationships among map entities. Graph-based encoders typically use the representations generated by token-based encoders as inputs and apply a graph neural network (GNN) to produce refined representation vectors. Specifically, given the map entity set $V = \{v_1, \ldots, v_I\}$, the embeddings generated by the token-based encoder are denoted as $\bm{R} = \{\bm{c}_1, \ldots, \bm{c}_I\}$. \updatem{The graph-based encoder then takes $\bm{R}$ along with the relation network $G$ to generate the final representations: $\bm{H} = \left\{\bm{h}_1, \ldots, \bm{h}_i, \ldots, \bm{h}_I\right\} = \mathrm{GNN}\left(\bm{R}, G\right),$ where $\bm{h}_i$ is the representation vector for the $i$-th map entity, and $\mathrm{GNN}(\cdot)$ is the GNN-based encoder model. Compared to token-based models, graph-based encoders incorporate the relation networks into the entity representations, enriching their expressiveness.}

Existing research on graph-based encoders focuses on developing GNNs capable of capturing complex relationships among map entities. Early studies applied standard GNN models, such as graph convolutional networks~\cite{stgcn,gwnet} and graph attention networks~\cite{STGAT,STGAT2}, to explore graphical~\cite{SARN,START,STPA} and social relations~\cite{MVURE,GMEL}. Given the diverse relationships among map entities, some studies adopt more compound GNNs to enhance graph-based encoders. For example, the HREP model constructs a heterogeneous graph with four types of relations to represent land parcels~\cite{HREP}. The MGFN model utilizes human mobility patterns to build a multi-graph~\cite{MGFN}. HRNR adopts a hierarchical GNN to capture the connection, structure, and function relations of road segments~\cite{HRNR}. \updateo{Recent advancements also investigate relationships across different types of map entities. For example, HGI~\cite{hgi} and HRoad~\cite{HyperRoad} utilize hypergraphs to capture these inter-entity relationships.}


\paratitle{Pre-training Tasks.} For graph-based models, the corresponding pre-training tasks include four types:

$\bullet$ {\em Node Feature Inference (NFI).} This task aims to predict the features of map entities based on the representations generated by graph-based encoders. Given an entity representation $\bm{h}_i$, the NFI task predicts its original features as $\hat{\bm{f}}_i = \phi(\bm{h}_i)$, where $\phi(\cdot)$ is typically an MLP-based prediction function. The objective is to minimize the prediction error.

$\bullet$ {\em Graph Autoencoder (GAu).} This task reconstructs the adjacency matrix of relation networks using an autoencoder framework. Given entity representations $\bm{H}$, the adjacency matrix is reconstructed as $\hat{\bm{E}} = \phi(\bm{H} \bm{H}^\top)$, where $\phi(\cdot)$ acts as a neural decoder. The graph encoder serves as the encoder of the auto-encoder, and the model is optimized to minimize the reconstruction error between $\hat{\bm{E}}$ and the original adjacency matrix $\bm{E}$.

$\bullet$ {\em Neighborhood Contrastive Learning (NCL).} The intuition behind this task is that connected map entities in a relation network should exhibit similar representations. This task treats connected entities as positive samples and unconnected entities as negative samples. Contrastive learning loss functions are then employed to optimize the encoder parameters.

\updatem{
$\bullet$ {\em Augmented Graph Contrastive Learning (AGCL).} This task introduces data augmentation to contrastive learning in relation networks. Given a relation network $G$ and its augmented variant $\tilde{G}$, the representations of the same node across these versions are treated as positive samples, while different nodes serve as negative samples.

In NFI tasks, most methods leverage raw entity features as supervision signals~\cite{HyperRoad,HRNR,USPM}. In particular, GMEL~\cite{GMEL} converts the OD flow relation network as labels for prediction. In the GAu task, the design of this task is closely linked to the structure of the GNN encoder. For example, HRNR~\cite{HRNR} uses a three-layer hierarchical GNN, requiring different adjacency matrix reconstruction tasks per layer. HRoad~\cite{HyperRoad} adopts a hypergraph in its encoder, so its pre-training step incorporates a hyperedge reconstruction task. In NCL tasks, the primary differences between models lie in the relation networks they utilize. For instance, HREP~\cite{HREP} and ReDCL~\cite{RegionDCL} connect the nearest entities to incorporate spatial information. In AGCL tasks, the choice of data augmentation is critical. Randomly deleting edges can disrupt essential connections. To mitigate this, some methods use edge weights to determine deletion probabilities~\cite{SARN,Garner}.}

\begin{table*}[t]
  \centering
  \vspace{-0.3cm}
  \caption{\updateo{Summary of existing \rl models. "*" indicates the reproduced models in \fullname.}}
  \vspace{-0.3cm}
  \resizebox{0.95\textwidth}{!}{

\begin{tabular}{c|r|c|ccc|ccc|cccc|ccc|ccc|rcr|cc}
    \toprule
    \multicolumn{2}{c|}{\multirow{2}{*}{Method}} & \textcolor{black}{Taxnomy} & \multicolumn{3}{c|}{Encoders} & \multicolumn{10}{c|}{Pretraining Tasks}                                       & \multicolumn{3}{c|}{Map Entity} & \multicolumn{3}{c|}{Features} & \multicolumn{2}{c}{Aux. Data} \\
\cline{4-24}    \multicolumn{2}{c|}{} & \textcolor{black}{Category}  & Token & Graph & Seq   & TokRI & TRCL  & AToCL & NFI   & GAu   & NCL   & AGCL  & TrajP & MTR   & ATrCL & POI   & RS    & LP    & \multicolumn{1}{c}{Sem} & \multicolumn{1}{c}{Spa} & \multicolumn{1}{c|}{Temp} & Traj  & Rel \\
    \midrule
    \multirow{13}[2]{*}{\rotatebox[origin=c]{90}{\textbf{Point of Interest}}} 
    & CWAP~\cite{CWAP}  & \textcolor{black}{Token-based} & \checkmark     &       &       &       & \checkmark     &       &       &       &       &       &       &       &       & \checkmark     &       &       &       & \multicolumn{1}{r}{} & \multicolumn{1}{c|}{\checkmark} & CIT   &  \\
    & CAPE~\cite{CAPE}  & \textcolor{black}{Token-based} & \checkmark     &       &       &       & \checkmark     &       &       &       &       &       &       &       &       & \checkmark     &       &       & \multicolumn{1}{c}{\checkmark} & \multicolumn{1}{r}{} &       & CIT   &  \\
    & \textcolor{black}{MT-POI}~\cite{MT-POI} & \textcolor{black}{Token-based} & \textcolor{black}{\checkmark}     &       &       &       & \textcolor{black}{\checkmark}     &       &       &       &       &       &       &       &       & \textcolor{black}{\checkmark}     &       &       & \multicolumn{1}{c}{\textcolor{black}{\checkmark}} & \multicolumn{1}{c}{\textcolor{black}{\checkmark}} &       &       & \textcolor{black}{GR} \\

    & \textcolor{black}{Place2V}~\cite{Place2Vec} & \textcolor{black}{Token-based} & \textcolor{black}{\checkmark} &  &  &  & \textcolor{black}{\checkmark} &  &  &  &  &  &  &  &  & \textcolor{black}{\checkmark} &  &  & \multicolumn{1}{c}{\textcolor{black}{\checkmark}} & \multicolumn{1}{l}{\textcolor{black}{\checkmark}} &  &  & \textcolor{black}{GR} \\
    & \textcolor{black}{Semant}~\cite{semantic} & \textcolor{black}{Token-based} & \textcolor{black}{\checkmark} &  &  &  & \textcolor{black}{\checkmark} &  &  &  &  &  &  &  &  & \textcolor{black}{\checkmark} &  &  & \multicolumn{1}{c}{\textcolor{black}{\checkmark}} & \multicolumn{1}{l}{\textcolor{black}{\checkmark}} &  &  & \textcolor{black}{GR} \\
    & \textcolor{black}{CatEM}~\cite{CatEM} & \textcolor{black}{Token-based} & \textcolor{black}{\checkmark} &  &  &  & \textcolor{black}{\checkmark} &  &  &  &  &  &  &  &  & \textcolor{black}{\checkmark} &  &  & \multicolumn{1}{c}{\textcolor{black}{\checkmark}} & \multicolumn{1}{l}{\textcolor{black}{\checkmark}} &  & \textcolor{black}{CIT}   &  \\
    
    & SkipG*~\cite{Word2Vec} & \textcolor{black}{Token-based} & \checkmark     &       &       &       & \checkmark     &       &       &       &       &       &       &       &       & \checkmark     &       &       &       & \multicolumn{1}{r}{} &       & CIT   &  \\
    & Tale*~\cite{TALE} & \textcolor{black}{Token-based} & \checkmark     &       &       & \checkmark     &       &       &       &       &       &       &       &       &       & \checkmark     &       &       &       & \multicolumn{1}{r}{} & \multicolumn{1}{c|}{\checkmark} & CIT   &  \\
    & Teaser*~\cite{Teaser} & \textcolor{black}{Token-based} & \checkmark     &       &       &       & \checkmark     &       &       &       &       &       &       &       &       & \checkmark     &       &       &       & \checkmark     & \multicolumn{1}{c|}{\checkmark} & CIT   & GR \\
    & Hier*~\cite{Hier1} & \textcolor{black}{Seq.-based} & \checkmark     &       & \checkmark     &       &       &       &       &       &       &       & \checkmark     &       &       & \checkmark     &       &       &       & \checkmark     &       & CIT   & GR \\
    & P2Vec*~\cite{POI2VEC} & \textcolor{black}{Token-based} & \checkmark     &       &       & \checkmark     &       &       &       &       &       &       &       &       &       & \checkmark     &       &       &       & \checkmark     &       & CIT   & GR \\
    & CACSR*~\cite{CACSR} & \textcolor{black}{Seq.-based} & \checkmark     &       & \checkmark     &       &       &       &       &       &       &       &       &       & \checkmark     & \checkmark     &       &       &       & \multicolumn{1}{c}{} & \multicolumn{1}{c|}{\checkmark} & CIT   &  \\
    & CTLE*~\cite{CTLE} & \textcolor{black}{Seq.-based} & \checkmark     &       & \checkmark     &       &       &       &       &       &       &       & \checkmark     & \checkmark     &       & \checkmark     &       &       &       & \checkmark     & \multicolumn{1}{c|}{\checkmark} & CIT   &  \\
    \midrule
    \multirow{11}[2]{*}{\rotatebox[origin=c]{90}{\textbf{Road Segment}}} 
    & DyToast~\cite{DyToast} & \textcolor{black}{Graph-based} & \checkmark     & \checkmark     &       &       & \checkmark     &       & \checkmark     &       &       &       &       & \checkmark     & \checkmark     &       & \checkmark     &       &       & \multicolumn{1}{l}{\checkmark} &       & CDT   & GR \\
    
    & \textcolor{black}{R2Vec*}~\cite{r2vec} & \textcolor{black}{Token-based} & \textcolor{black}{\checkmark} &  &  &  & \textcolor{black}{\checkmark} &  &  &  &  &  &  &  &  &  & \textcolor{black}{\checkmark} &  &  & \textcolor{black}{\checkmark} &  &  & \textcolor{black}{GR} \\
    & \textcolor{black}{TrajRNE}~\cite{TrajRNE} & \textcolor{black}{Graph-based} & \textcolor{black}{\checkmark} & \textcolor{black}{\checkmark} &  & \textcolor{black}{\checkmark} &  &  &  & \textcolor{black}{\checkmark} &  &  &  &  &  &  & \textcolor{black}{\checkmark} &  & \multicolumn{1}{c}{\textcolor{black}{\checkmark}} &  &  & \textcolor{black}{CDT}   & \textcolor{black}{GR/SR} \\
    & \textcolor{black}{JGRM}~\cite{JGRM} & \textcolor{black}{Seq.-based} & \textcolor{black}{\checkmark} & \textcolor{black}{\checkmark} & \textcolor{black}{\checkmark} &  &  &  &  &  &  &  &  & \textcolor{black}{\checkmark} & \textcolor{black}{\checkmark} &  & \textcolor{black}{\checkmark} &  & \multicolumn{1}{c}{\textcolor{black}{\checkmark}} & \textcolor{black}{\checkmark} & \multicolumn{1}{c|}{\textcolor{black}{\checkmark}} & \textcolor{black}{CDT}   & \textcolor{black}{GR} \\

    & RN2Vec*~\cite{RN2Vec} & \textcolor{black}{Token-based} & \checkmark     &       &       & \checkmark     &       &       &       &       &       &       &       &       &       &       & \checkmark     &       & \multicolumn{1}{c}{\checkmark} & \checkmark     &       &       & GR \\
    & HRNR*~\cite{HRNR} & \textcolor{black}{Graph-based} & \checkmark     & \checkmark     &       &       &       &       &       & \checkmark     &       &       &       &       &       &       & \checkmark     &       & \multicolumn{1}{c}{\checkmark} & \checkmark     &       &       & GR \\
    & SARN*~\cite{SARN} & \textcolor{black}{Graph-based} & \checkmark     & \checkmark     &       &       &       &       &       &       &       & \checkmark     &       &       &       &       & \checkmark     &       & \multicolumn{1}{c}{\checkmark} & \checkmark     &       &       & GR \\
    & Toast*~\cite{Toast} & \textcolor{black}{Seq.-based} & \checkmark     &       & \checkmark     &       & \checkmark     &       &       &       &       &       &       & \checkmark     & \checkmark     &       & \checkmark     &       &       & \multicolumn{1}{r}{} & \multicolumn{1}{c|}{\checkmark} & CDT   & GR \\
    & HRoad*~\cite{HyperRoad} & \textcolor{black}{Graph-based} & \checkmark     & \checkmark     &       &       &       &       & \checkmark     & \checkmark     &       &       &       &       &       &       & \checkmark     &       & \multicolumn{1}{c}{\checkmark} & \checkmark     &       &       & GR \\
    & START*~\cite{START} & \textcolor{black}{Seq.-based} & \checkmark     & \checkmark     & \checkmark     &       &       &       &       &       &       &       &       & \checkmark     & \checkmark     &       & \checkmark     &       & \multicolumn{1}{c}{\checkmark} & \checkmark     & \multicolumn{1}{c|}{\checkmark} & CDT   & GR \\
    & JCLRNT*~\cite{JCLRNT} & \textcolor{black}{Seq.-based} & \checkmark     & \checkmark     & \checkmark     &       &       &       &       &       &       & \checkmark     &       &       & \checkmark     &       & \checkmark     &       & \multicolumn{1}{c}{\checkmark} & \checkmark     & \multicolumn{1}{c|}{\checkmark} & CDT   & GR \\
    \midrule
    \multirow{15}[2]{*}{\rotatebox[origin=c]{90}{\textbf{Land Parcel}}} 
    & CGAL~\cite{CGAL}  & \textcolor{black}{Graph-based} & \checkmark     & \checkmark     &       &       &       &       &       & \checkmark     &       &       &       &       &       & \checkmark     &       & \checkmark     & \multicolumn{1}{c}{\checkmark} & \checkmark     &       &       & GR/SR \\
    & DLCL~\cite{DLCL}  & \textcolor{black}{Graph-based} & \checkmark     & \checkmark     &       &       &       &       &       & \checkmark     &       &       &       &       &       & \checkmark     &       & \checkmark     & \multicolumn{1}{c}{\checkmark} & \checkmark     &       &       & GR/SR \\
    & HUGAT~\cite{HUGAT} & \textcolor{black}{Graph-based} & \checkmark     & \checkmark     &       &       &       &       &       & \checkmark     &       &       &       &       &       &       &       & \checkmark     & \multicolumn{1}{c}{\checkmark} & \checkmark     & \multicolumn{1}{c|}{\checkmark} &       & GR/SR \\
    & Re2Vec~\cite{Region2Vec} & \textcolor{black}{Graph-based} & \checkmark     & \checkmark     &       &       &       &       &       & \checkmark     &       &       &       &       &       & \checkmark     &       & \checkmark     & \multicolumn{1}{c}{\checkmark} & \checkmark     &       &       & GR/SR \\
    & ReCP~\cite{ReCP}  & \textcolor{black}{Token-based} & \checkmark     &       &       &  \checkmark & \checkmark     & \checkmark  &      &       &       &       &       &       &   & \checkmark     &       & \checkmark     & \multicolumn{1}{c}{\checkmark} & \checkmark     &       &       & SR \\

    & \textcolor{black}{HGI}~\cite{hgi}  & \textcolor{black}{Graph-based} & \textcolor{black}{\checkmark} & \textcolor{black}{\checkmark} &  &  & \textcolor{black}{\checkmark} & \textcolor{black}{\checkmark} &  &  &  &  &  &  &  & \textcolor{black}{\checkmark} &  & \textcolor{black}{\checkmark} & \multicolumn{1}{c}{\textcolor{black}{\checkmark}} & \textcolor{black}{\checkmark} &  &  & \textcolor{black}{GR} \\
    & \textcolor{black}{ReDCL}~\cite{RegionDCL} & \textcolor{black}{Token-based} & \textcolor{black}{\checkmark} &  &  & \textcolor{black}{\checkmark} & \textcolor{black}{\checkmark} & \textcolor{black}{\checkmark} &  &  &  &  &  &  &  & \textcolor{black}{\checkmark} &  & \textcolor{black}{\checkmark} & \multicolumn{1}{c}{\textcolor{black}{\checkmark}} & \textcolor{black}{\checkmark} &  &  & \textcolor{black}{GR} \\
    & \textcolor{black}{HAFus}~\cite{HAFusion} & \textcolor{black}{Graph-based} & \textcolor{black}{\checkmark} & \textcolor{black}{\checkmark} &  &  &  &  &  & \textcolor{black}{\checkmark} &  &  &  &  &  & \textcolor{black}{\checkmark} &  & \textcolor{black}{\checkmark} & \multicolumn{1}{c}{\textcolor{black}{\checkmark}} & \textcolor{black}{\checkmark} &  &  & \textcolor{black}{GR/SR} \\

    & ZEMob*~\cite{ZEMob} & \textcolor{black}{Token-based} & \checkmark     &       &       &       & \checkmark     &       &       &       &       &       &       &       &       &       &       & \checkmark     &       & \multicolumn{1}{r}{} & \multicolumn{1}{c|}{\checkmark} &       & SR \\
    & GMEL*~\cite{GMEL} & \textcolor{black}{Graph-based} & \checkmark     & \checkmark     &       &       &       &       & \checkmark     & \checkmark     &       &       &       &       &       &       &       & \checkmark     &       & \checkmark     &       &       & GR/SR \\
    & MGFN*~\cite{MGFN} & \textcolor{black}{Graph-based} & \checkmark     & \checkmark     &       &       &       &       &       & \checkmark     &       &       &       &       &       &       &       & \checkmark     &       & \checkmark     &       &       & SR \\
    & HDGE*~\cite{HDGE} & \textcolor{black}{Token-based} & \checkmark     &       &       &       & \checkmark     &       &       &       &       &       &       &       &       &       &       & \checkmark     &       & \checkmark     & \multicolumn{1}{c|}{\checkmark} &       & GR/SR \\
    & MVURE*~\cite{MVURE} & \textcolor{black}{Graph-based} & \checkmark     & \checkmark     &       &       &       &       &       & \checkmark     &       &       &       &       &       & \checkmark     &       & \checkmark     &       & \checkmark     &       &       & SR \\
    & ReMVC*~\cite{REMVC} & \textcolor{black}{Token-based} & \checkmark     &       &       &       &       & \checkmark     &       &       &       &       &       &       &       & \checkmark     &       & \checkmark     & \multicolumn{1}{c}{\checkmark} & \checkmark     &       &       & SR \\
    & HREP*~\cite{HREP} & \textcolor{black}{Graph-based} & \checkmark     & \checkmark     &       &       &       &       &       & \checkmark     & \checkmark     &       &       &       &       & \checkmark     &       & \checkmark     &       & \checkmark     &       &       & GR/SR \\
    \bottomrule
    \end{tabular}%
        }
  \label{tab:models}%
    \vspace{-0.4cm}
\end{table*}%

\subsubsection{Sequence-based Models}\;

\paratitle{Encoders.} \updatem{Sequence-based encoders are designed to capture temporal dependencies among map entities within trajectory-type auxiliary data. They employ sequential deep learning models, such as LSTMs~\cite{LSTM}, GRUs~\cite{GRU}, or Transformers~\cite{attention}, to encode the embeddings of map entities along a trajectory into a sequence of representation vectors. These input embeddings are typically derived from token-based or graph-based encoders. Specifically, for a trajectory $tr$ consisting of 
$K$ entities, let their embedding sequence be {\small $\bm{R}_{tr} = (\bm{r}_{tr_1}, \ldots, \bm{r}_{tr_k}, \ldots, \bm{r}_{tr_K})$}. The sequence-based model then processes this input to generate corresponding representation vectors as $\bm{S}_{tr} = (\bm{s}_{tr_1}, \ldots, \bm{s}_{tr_k}, \ldots, \bm{s}_{tr_K}) = \mathrm{SEQ}(\bm{R}_{tr}),$ where $\mathrm{SEQ}(\cdot)$ denotes the sequence-based encoder, and $\bm{s}_{tr_k}$ represents the embedding of the $k$-th entity along the trajectory $tr$.}

In trajectory data, each location sample is associated with a timestamp. Integrating temporal information from timestamps into entity representations has become a key focus of sequence-based encoders. Early methods divide time into discrete slots, such as the hour of the day or the day of the week, and encoding these slots as embeddings using token-based encoders~\cite{JCLRNT,POI2VEC,Teaser}. More advanced models embed timestamps within the components of sequential models. For example, JCLRNT~\cite{JCLRNT}, Toast~\cite{Toast}, and CTLE~\cite{CTLE} incorporate timestamps into the positional encoding of Transformers. Similarly, START~\cite{START} integrates travel time into the attention mechanism, enhancing the model's ability to capture temporal dependencies.

\paratitle{Pre-training Tasks.} For sequence-based models, the corresponding pre-training tasks include three types:

$\bullet$ {\em Trajectory Prediction (TrajP).} The goal of this task is to predict the second half of a trajectory using the first half. Given a trajectory $tr$, this task is formally defined as $\hat{tr}_{[k+1:K]} = Pre(\bm{S}_{[1:k]}), ~\bm{S}_{[1:k]} = Enc(tr_{[1:k]})$, where $Enc(\cdot)$ is the sequence-based encoder, and $Pre(\cdot)$ is the prediction function. Here, {\small $tr_{[1:k]}$} represents the first $k$ samples of the trajectory, and {\small $\bm{S}_{[1:k]}$} is their corresponding sequence of representation vectors. The objective is to minimize the distance between $\hat{tr}_{[k+1:K]}$ and the ground truth.

\updatem{
$\bullet${\em Masked Trajectory Recovery (MTR).} Inspired by the masked language model (MLM) in NLP~\cite{bert,roberta}, this task randomly masks parts of a trajectory and trains an autoencoder for reconstruction. Given a trajectory $tr$ and its masked version $\tilde{tr}$, the recovery process follows $\hat{tr}= Dec(\tilde{\bm{S}}),~ \tilde{\bm{S}}=Enc(\tilde{tr})$, where $Enc(\cdot)$ is the sequence-based encoder, and $Dec(\cdot)$ is the decoder. The objective is to minimize the reconstruction error between $tr$ and $\hat{tr}$.

$\bullet$ {\em Augmented Trajectory Contrastive Learning (ATrCL).} This task generates multiple versions of a trajectory using data augmentation techniques~\cite{tra1,tra2}. Sequence-based encoders then encode both the original and augmented trajectories, and a contrastive learning loss is applied to enhance the similarity between representations of the same trajectory while distinguishing them from others.}

In TrajP tasks, the prediction labels can be various. For example, Hier~\cite{Hier1} uses entity IDs as prediction labels, while CTLE~\cite{CTLE} employs visiting time as labels to incorporate temporal information. In MTR tasks, different strategies exist for selecting masked positions. A basic approach is to mask positions within trajectories~\cite{CTLE} randomly, but this overlooks the sequential dependencies between samples. To address this, START~\cite{START} and Toast~\cite{Toast} mask continuous subsequences within a trajectory. In ATrCL tasks, models adopt diverse augmentation strategies. JCLRNT~\cite{JCLRNT} generates augmented trajectories by randomly deleting or replacing trajectory points. Considering contextual dependencies, START~\cite{START} replaces continuous sub-sequences, while CACSR~\cite{CACSR} injects white noise into representations to enhance robustness.

\vspace{-2mm}
\subsubsection{Module Integration of Encoders and Pre-training Tasks}\;
\vspace{1mm}

The encoders and pre-training tasks introduced in the previous sections serve as the foundational components of \rl models. A complete \rl model can integrate multiple types of encoders and pre-training tasks, leveraging their complementary strengths to enhance representation learning. For the encoder modules, existing models typically follow a ``token $\rightarrow$ graph $\rightarrow$ sequence'' modeling pipeline. This pipeline first extracts basic features using token-based encoders, then captures relational structures through graph-based encoders, and finally models temporal dependencies with sequence-based encoders. At each stage of this pipeline, a \rl model may employ multiple encoders simultaneously for different pre-training tasks. Table~\ref{tab:models} provides an overview of how mainstream models combine encoders and pre-training tasks.

For the three-stage encoder pipeline, \rl models follow two pre-training paradigms. $i$) {\em Sequential Training}: Encoders are trained in stages—first token-based, then graph-based, and finally sequence-based—where each stage refines the output of the previous one. $ii$) {\em Joint Training}: All encoders are optimized simultaneously using multi-task learning, enabling the model to capture entity features, structural relationships, and temporal patterns in an integrated manner. These paradigms offer flexibility in model impanelments, depending on the specific requirements of the \rl task.

\vspace{-0.1cm}
\subsection{Downstream Tasks}~\label{sec:taxonomy_downstream}
\vspace{-0.1cm}

The downstream tasks consist of two components: {\em Downstream Models} and {\em Fine-tuning Loss}. Downstream models use the representation vectors generated by the encoders to perform specific tasks. Fine-tuning loss adjusts the encoder parameters to improve the adaptability of the representation vectors for these tasks. In contrast to pre-training tasks, which are closely tied to the encoders, downstream tasks are heavily influenced by the type of map entities. This subsection presents a taxonomy of downstream tasks.

\paratitle{Tasks for POIs.} Common downstream tasks for POI entities can be classified into three categories, all of which utilize cross-entropy loss as the objective function:

$\bullet$ {\em POI Classification (POIC)}: This task aims to classify a POI into predefined categories based on its representation. A typical downstream model consists of an MLP followed by a SoftMax classifier.

$\bullet$ {\em Next POI Prediction (NPP)}: This task converts a POI-based trajectory into a sequence of representation vectors using an encoder. The downstream model, often a deep sequential model like LSTM, predicts the next POI in the trajectory using the representation of the first half. The output is a multi-class classification, with each class representing a possible POI.

$\bullet$ {\em Trajectory User Link (TUL)}: This task predicts the user who generated a trajectory based on the trajectory's representation sequence. The downstream model is a deep sequential model, with each output class corresponding to a user ID.


\paratitle{Tasks for Road Segments.} Common downstream tasks for road segment entities fall into three categories:

$\bullet$ {\em Average Speed Inference \textit{(ASI)}.} This task leverages the representation vector to infer the average speed of a road segment. The average speed is derived from trajectory data and must be excluded from the input features of the representation encoders to prevent data leakage. Inferring average speed is a representative task for estimating unknown features of a road segment. The downstream model for this task is a linear regression function, and the fine-tuning loss function is Mean Square Error (MSE).

$\bullet$ {\em Travel Time Estimation (TTE).} This task estimates the travel time from the origin to the destination along a trajectory composed of road segments. The downstream model is a deep sequential model. Its input is the representation sequence of the segment-based trajectory, and the output is the regressed travel time. The fine-tuning loss function is MSE.

\updatem{
$\bullet$ {\em Similarity Trajectory Search (STS).} 
This task retrieves the most similar trajectory from a database given a query. A deep sequential model encodes all trajectories into representation vectors, and cosine similarity identifies the closest match. Fine-tuning uses a contrastive loss (e.g., InfoNCE), treating the ground-truth trajectory—formed by replacing a sub-trajectory with one sharing the same endpoints~\cite{START,JCLRNT,SARN}—as the positive sample.
}

\paratitle{Tasks for Land Parcels.}
The downstream tasks for land parcel entities primarily focus on attribute classification, such as function and population, and mobility volume. These tasks are typically categorized into three types:

$\bullet$ {\em Land Parcel Classification (LPC)}: This task predicts specific labels for a land parcel, such as land usage or functional category, based on its representation vector. The downstream model is an MLP with a SoftMax classifier, optimized with cross-entropy loss.

$\bullet$ {\em Flow Inference (FI)}: This task predicts the incoming and outgoing population of a land parcel over a given period (\eg one month) using its representation vector. The downstream function is usually a linear regression, with MSE as the loss.

$\bullet$ {\em Mobility Inference (MI)}: This task predicts the OD flow between pairs of land parcels. The input to the downstream function consists of the representations of two parcels, and the output is the predicted OD flow volume. The downstream function can be a neural network, with MSE as the fine-tuning loss.

There are two fine-tuning strategies for downstream tasks~\cite{SARN,T5}. $i$) {\em Downstream Fine-tuning}: In this strategy, only the downstream models are fine-tuned, while the parameters of the \rl encoders remain frozen. $ii$) {\em End-to-End Fine-tuning}: This approach tunes the parameters of both the downstream models and the \rl encoders, enabling the entire model to be optimized jointly.

\section{The Library -- \name}~\label{sec:library}

\begin{figure}[ht]
    \centering
    \includegraphics[width=0.8\linewidth]{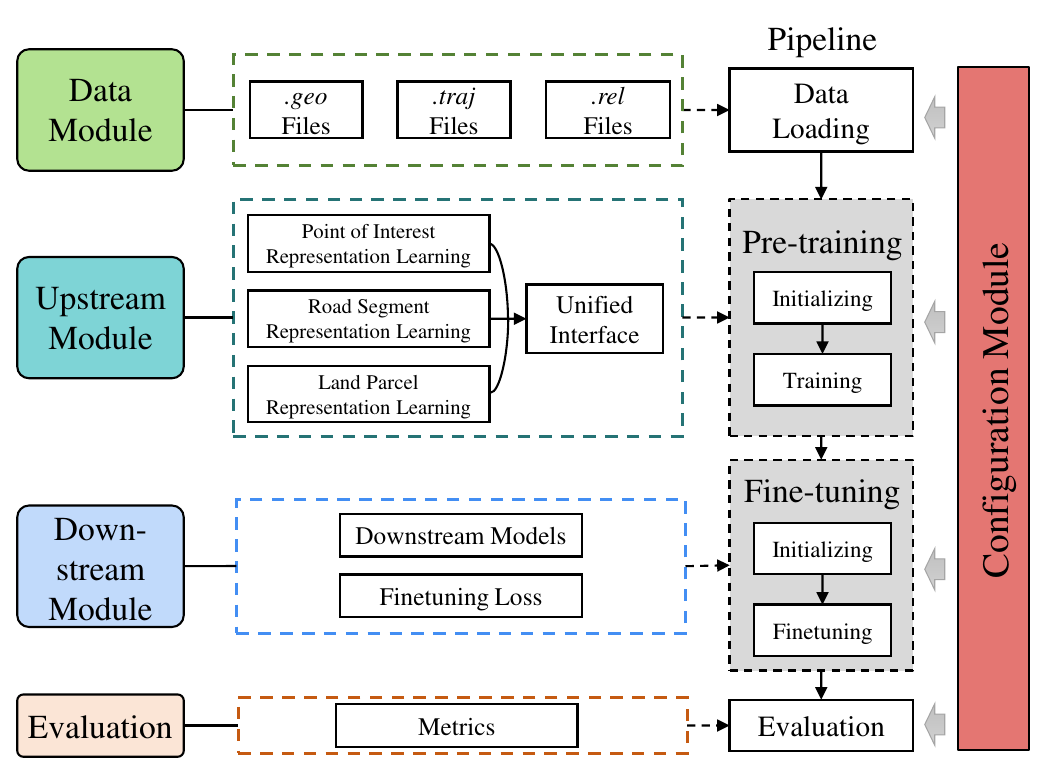}
    \vspace{-0.2cm}
    \caption{The overall pipeline for \name.}
    \vspace{-0.4cm}
    \label{fig:framework}
\end{figure}

Based on the \tax of map entity representation learning models, we propose \name, an easy-to-use toolkit for \rl model development. Figure~\ref{fig:framework} illustrates the framework of \fullname, comprising three functional modules aligned with the four key elements of \rl models outlined in Section~\ref{sec:taxonomy}. $i$) {\em Data Module}: Corresponding to the map data element in Section~\ref{sec:taxonomy_data}, this module handles data preparation and processing. $ii$) {\em Upstream Module}: Aligned with the encoder and pre-training elements in Section~\ref{sec:taxonomy_pretraining}, it is responsible for representation encoding and model pre-training. $iii$) {\em Downstream Module}: Linked to the downstream task element in Section~\ref{sec:taxonomy_downstream}, this module manages model fine-tuning and performance evaluation on downstream tasks.

\subsection{Data Module and Atomic Files}

In the Data Module, map entities and auxiliary data are represented in a unified file format, called {\em Atomic Files}~\cite{libcity}. This format consists of three file types: {\em .geo} – represents map entities, {\em .traj} – stores trajectory data, {\em .rel} – captures relation networks. Using the three types of atomic files, we can represent map entities and auxiliary data in a unified format. Original data often comes in various spatiotemporal formats, such as GeoJSON~\cite{geojson}, Shapefile~\cite{shapefile}, and NPZ~\cite{harris2020array}, requiring multiple data-loading implementations and hindering reusability. Our \name library solves this issue by adopting unified atomic files, simplifying data processing.

In \name, we have pre-loaded map entities and auxiliary data from nine cities: New York (NY), Chicago (CHI), Tokyo (TYO), Singapore (SIN), Porto (PRT), San Francisco (SF), Beijing (BJ), Chengdu (CD), and Xi'an (XA), using public data sources. Map entities are extracted from OpenStreetMap (OSM), an open-source global map. POI check-in trajectories come from Foursquare datasets~\cite{fours1}, while coordinate trajectories are collected from various studies~\cite{START, didi2022gaia, porto, sanfransico}. The trajectory data preprocessing approach aligns with prior studies~\cite{deepmove}. Relation networks are OD flow networks constructed from these trajectory datasets. All datasets are processed, converted into atomic files, and integrated into the \name library. Details of the datasets are provided in Tab.~\ref{tab:datasets}. Users can directly apply these datasets in their model development. Additionally, we provide data conversion scripts on the \name GitHub repository, enabling users to convert their private data into the atomic file format.

\subsection{Upstream Module and Downstream Module}

\paratitle{Upstream Module.} Based on the taxonomy of \rl encoders and pre-training tasks in Section~\ref{sec:taxonomy_pretraining}, \fullname provides the upstream module through two key interfaces: {\em encode()} and {\em pretraining\_loss()}.

$\bullet$ {\em encode()}: This interface implements the encoder for an \rl model. It takes data extracted from atomic files as input and outputs representation vectors for map entities.

$\bullet$ {\em pretraining\_loss()}: This interface manages the implementation of pre-training tasks. It prepares samples through methods such as data masking, generating positive/negative samples, or data augmentation, and calculates the corresponding loss values.

In the \name framework, users can execute the pre-training phase of an \rl model by implementing the interface functions in the upstream module. Leveraging these interfaces, we have reproduced 21 mainstream \rl models in \name, including 7 models each for POI, road segment, and land parcel representation learning. The reproduced models are marked with ``*'' in Tab.~\ref{tab:models}.

\paratitle{Downstream Module.} The downstream module provides two key interfaces: {\em downstream\_model()} and {\em finetuning\_loss()}.

$\bullet$ {\em downstream\_model()}: This interface implements the downstream models used during the fine-tuning step. It typically utilizes a simple MLP or LSTM model. For classification tasks, the output layer is a SoftMax classifier, while for continuous prediction tasks, it is a linear regression layer.

$\bullet$ {\em finetuning\_loss()}: This interface defines the loss function for fine-tuning. Cross-entropy is used for classification tasks, MSE for regression tasks, and InfoNCE for contrastive learning tasks.

Using these two interfaces, we have implemented all nine downstream tasks listed in Section~\ref{sec:taxonomy_downstream}. Users can directly invoke these pre-set tasks to fine-tune their models.

\begin{table}[t]\small
  \centering
  \vspace{-0.3cm}
  \caption{Pre-prepared datasets for \fullname. }
  \vspace{-0.3cm}
  \setlength{\tabcolsep}{0.8mm}
  \resizebox{\columnwidth}{!}{
    \begin{tabular}{l|rrr|rr|rr|r}
    \toprule
    \multirow{2}{*}{City} & \multirow{2}{*}{\#POI} & \multicolumn{1}{c}{\multirow{2}{*}{\#Segment}} & \multicolumn{1}{c|}{\multirow{2}{*}{\#Parcel}} & \multicolumn{2}{c|}{Check-in Traj} & \multicolumn{2}{c|}{Coor. Traj} & \multicolumn{1}{c}{\multirow{2}{*}{\#OD}} \\
\cline{5-8}          &       &       &       & \multicolumn{1}{c}{\#Traj} & \multicolumn{1}{c|}{\#User} & \multicolumn{1}{c}{\#Traj} & \multicolumn{1}{c|}{\#User} &  \\
    \midrule
    NY    & 79,627 & 90,781 & 262   & 823,853 & 2,758 & \multicolumn{1}{c}{-}     & \multicolumn{1}{c|}{-}     & 28,189 \\
    CHI   & 28,141 & 47,669 & 77    & 279,183 & 921   & \multicolumn{1}{c}{-}    & \multicolumn{1}{c|}{-}     & 4,631 \\
    TYO   & 61,858 & 407,905 & 64    & 573,703 & 2,227 & 226,782 & \multicolumn{1}{c|}{-}  &\multicolumn{1}{c}{-} \\
    SIN   & 75,329 & 35,084 & 332   & 827,828 & 3,489 & 11,170 & \multicolumn{1}{c|}{-}  & \multicolumn{1}{c}{-} \\
    PRT   & 4,521 & 11,095 & 382   & 19,402 & 126   & 695,085 & 435   & 324 \\
    SF    & 15,674 & 27,274 & 194   & 171,873 & 361   & 500,516 & 405   & 24,716 \\
    BJ    & 81,181 & 40,306 & 11,208 & \multicolumn{1}{c}{-}    & \multicolumn{1}{c|}{-}     & 1,018,312 & 1,677 & 687,890 \\
    CD    & 17,301 & 6,195 & 1,306 & \multicolumn{1}{c}{-}     & \multicolumn{1}{c|}{-}     & 559,729 & 48,295 & 64,482 \\
    XA    & 19,108 & 5,269 & 1,056 & \multicolumn{1}{c}{-}     & \multicolumn{1}{c|}{-}     & 384,618 & 26,787 & 54,365 \\
    \bottomrule
    \end{tabular}%
    }
  \label{tab:datasets}%
  \vspace{-0.6cm}
\end{table}%

\subsection{Evaluation \& Config Modules}

The evaluation module provides an {\em evaluation()} interface to assess \rl model performance across downstream tasks. \fullname supports various metrics tailored to different task types: MAE, MSE, RMSE, R$^2$, and MAPE for regression tasks; Accuracy@k, Recall@k, F1@k, and Mean Rank (MR) for classification tasks; and Accuracy@k and MR for trajectory similarity search.

The config module serves as the control center, allowing users to configure the entire ``pre-training $\rightarrow$ fine-tuning $\rightarrow$ evaluation'' pipeline through a configuration file. To train and evaluate a model with \name, users simply specify the required pre-built or custom models and functions in the configuration file. The library then automatically generates the necessary script to run experiments. Users only need to execute the script, and \name will invoke the interface functions of each module to complete the experiment.

\section{Performance Benchmark}\label{sec:experiment}

In this section, we present performance comparison experiments for the pre-built \rl models in \name. These experiments serve three purposes: $i$) Demonstrating Usability and Power: To show that \name is a user-friendly and powerful toolkit for \rl model development and evaluation. $ii$) Establishing Performance Benchmarks: To provide standard performance comparisons of mainstream \rl models using the datasets pre-loaded in \name. \updatetr{$iii$) Analyzing Model Efficiency. Evaluating the computational efficiency of different \rl models to offer insights for designing more efficient approaches.}

\subsection{Datasets and Experimental Setups}

\subsubsection{Datasets} The dataset selection for our experiments follows these guiding principles: for each type of \rl model, we incorporate all datasets that meet the task requirements. Specifically, for POI-oriented \rl models, we use datasets from New York, Chicago, Tokyo, Singapore, and San Francisco. For road-segment-oriented models, the datasets include Beijing, Chengdu, Xi'an, Porto, and San Francisco. For land-parcel-oriented models, we use datasets from Beijing, Chengdu, Xi'an, Porto, and San Francisco. Each dataset is divided into training, validation, and test sets following a 6:2:2 ratio, ensuring a balanced and reliable performance evaluation. For each downstream task, we apply end-to-end fine-tuning to achieve optimal performance.

\subsubsection{Experimental Settings} In our experiments, the dimension of map entity representation vectors across all models is set to 128. For other model configurations, such as the number of layers and the dimensions of intermediate variables, we adopt the default settings from the original papers as initial values. We then use a trial-and-error approach to tune the parameters on each dataset, searching for the best performance configurations. \updateo{Each model is run five times with different seeds, and the results are reported as the mean and standard deviation.} All experiments are executed on an Ubuntu 20.04 system equipped with NVIDIA GeForce 3090 GPUs. The default batch size during model training is set to 64. In cases of out-of-memory (OOM) issues, the batch size is halved iteratively, with a minimum batch size of 8.

\begin{table}[t]
  \centering
  \vspace{-0.3cm}
  \caption{\updatetw{Performance comparison of POI \rl methods across various datasets (Mean ± Std). Bold and underlined text indicates the best and second-best results, respectively. '↑' denotes that a higher value is better, while '↓' indicates that a lower value is preferred.}}
\vspace{-0.3cm}
    \setlength{\tabcolsep}{0.4mm}
  \resizebox{\linewidth}{!}{
    \begin{tabular}{ccc|ccccccc}
    \toprule
    \multicolumn{1}{c|}{Data} & \multicolumn{1}{c|}{Task} & Metric & SkipG & Tale  & Teaser & Hier  & P2Vec & CACSR & CTLE \\
    \midrule
    \multicolumn{1}{c|}{\multirow{6}[2]{*}{\rotatebox{90}{New York}}} & \multicolumn{1}{c|}{\multirow{2}[1]{*}{POIC}} & ACC@1↑ & \underline{0.07±0.00} & 0.06±0.01 & 0.07±0.01 & 0.07±0.01 & \textbf{0.08±0.01} & 0.06±0.02 & 0.06±0.01 \\
    \multicolumn{1}{c|}{} & \multicolumn{1}{c|}{} & F1↑   & 0.01±0.00 & 0.02±0.01 & \underline{0.01±0.01} & 0.01±0.00 & \textbf{0.02±0.01} & 0.01±0.00 & 0.01±0.00 \\
    \multicolumn{1}{c|}{} & \multicolumn{1}{c|}{\multirow{2}[0]{*}{NPP}} & ACC@1↑ & 0.10±0.02 & \underline{0.14±0.01} & 0.10±0.02 & 0.12±0.00 & 0.13±0.01 & 0.14±0.05 & \textbf{0.15±0.01} \\
    \multicolumn{1}{c|}{} & \multicolumn{1}{c|}{} & ACC@5↑ & 0.25±0.04 & 0.33±0.00 & 0.25±0.05 & 0.29±0.01 & 0.16±0.00 & \underline{0.33±0.04} & \textbf{0.33±0.01} \\
    \multicolumn{1}{c|}{} & \multicolumn{1}{c|}{\multirow{2}[1]{*}{TUL}} & ACC@1↑ & 0.57±0.03 & 0.67±0.00 & 0.57±0.02 & 0.58±0.03 & \underline{0.64±0.01} & 0.64±0.03 & \textbf{0.69±0.00} \\
    \multicolumn{1}{c|}{} & \multicolumn{1}{c|}{} & F1↑   & 0.26±0.03 & \underline{0.44±0.01} & 0.27±0.03 & 0.28±0.04 & 0.33±0.00 & 0.39±0.07 & \textbf{0.45±0.01} \\
    \midrule
    \multicolumn{1}{c|}{\multirow{6}[2]{*}{\rotatebox{90}{Tokyo}}} & \multicolumn{1}{c|}{\multirow{2}[1]{*}{POIC}} & ACC@1↑ & \underline{0.35±0.01} & 0.22±0.00 & \textbf{0.38±0.02} & 0.34±0.04 & 0.32±0.00 & 0.30±0.04 & 0.30±0.04 \\
    \multicolumn{1}{c|}{} & \multicolumn{1}{c|}{} & F1↑   & 0.02±0.01 & \textbf{0.05±0.01} & 0.02±0.01 & 0.01±0.01 & \underline{0.04±0.00} & 0.02±0.00 & 0.02±0.00 \\
    \multicolumn{1}{c|}{} & \multicolumn{1}{c|}{\multirow{2}[0]{*}{NPP}} & ACC@1↑ & 0.16±0.00 & 0.16±0.01 & 0.16±0.01 & \textbf{0.16±0.00} & \underline{0.16±0.01} & 0.16±0.06 & 0.16±0.00 \\
    \multicolumn{1}{c|}{} & \multicolumn{1}{c|}{} & ACC@5↑ & 0.32±0.01 & 0.30±0.00 & 0.30±0.04 & \textbf{0.34±0.00} & \underline{0.33±0.01} & 0.33±0.01 & 0.33±0.01 \\
    \multicolumn{1}{c|}{} & \multicolumn{1}{c|}{\multirow{2}[1]{*}{TUL}} & ACC@1↑ & 0.43±0.01 & 0.40±0.01 & 0.44±0.02 & 0.44±0.03 & 0.45±0.00 & \underline{0.46±0.10} & \textbf{0.50±0.00} \\
    \multicolumn{1}{c|}{} & \multicolumn{1}{c|}{} & F1↑   & 0.23±0.01 & 0.23±0.01 & 0.23±0.01 & 0.23±0.01 & 0.25±0.00 & \underline{0.28±0.10} & \textbf{0.29±0.00} \\
    \midrule
    \multicolumn{1}{c|}{\multirow{6}[2]{*}{\rotatebox{90}{Chicago}}} & \multicolumn{1}{c|}{\multirow{2}[1]{*}{POIC}} & ACC@1↑ & 0.07±0.00 & 0.10±0.00 & 0.07±0.01 & 0.08±0.00 & \textbf{0.12±0.01} & \underline{0.10±0.01} & 0.10±0.02 \\
    \multicolumn{1}{c|}{} & \multicolumn{1}{c|}{} & F1↑   & 0.01±0.00 & 0.02±0.01 & 0.00±0.00 & 0.01±0.00 & \textbf{0.02±0.01} & 0.02±0.01 & \underline{0.02±0.01} \\
    \multicolumn{1}{c|}{} & \multicolumn{1}{c|}{\multirow{2}[0]{*}{NPP}} & ACC@1↑ & 0.08±0.02 & 0.10±0.01 & 0.06±0.03 & 0.10±0.03 & 0.15±0.01 & \underline{0.15±0.05} & \textbf{0.18±0.01} \\
    \multicolumn{1}{c|}{} & \multicolumn{1}{c|}{} & ACC@5↑ & 0.24±0.05 & 0.24±0.01 & 0.18±0.08 & 0.26±0.06 & 0.36±0.01 & \textbf{0.40±0.04} & \underline{0.38±0.01} \\
    \multicolumn{1}{c|}{} & \multicolumn{1}{c|}{\multirow{2}[1]{*}{TUL}} & ACC@1↑ & 0.46±0.08 & 0.62±0.01 & 0.43±0.07 & 0.55±0.02 & 0.66±0.01 & \underline{0.67±0.07} & \textbf{0.76±0.01} \\
    \multicolumn{1}{c|}{} & \multicolumn{1}{c|}{} & F1↑   & 0.20±0.07 & 0.37±0.01 & 0.19±0.01 & 0.30±0.01 & 0.42±0.01 & \underline{0.44±0.06} & \textbf{0.60±0.03} \\
    \midrule
    \multicolumn{1}{c|}{\multirow{6}[2]{*}{\rotatebox{90}{Singapore}}} & \multicolumn{1}{c|}{\multirow{2}[1]{*}{POIC}} & ACC@1↑ & 0.09±0.04 & 0.04±0.01 & 0.06±0.00 & \underline{0.12±0.04} & 0.07±0.00 & 0.10±0.04 & \textbf{0.13±0.03} \\
    \multicolumn{1}{c|}{} & \multicolumn{1}{c|}{} & F1↑   & 0.00±0.00 & 0.01±0.01 & \underline{0.01±0.01} & 0.00±0.00 & \textbf{0.03±0.01} & 0.01±0.00 & 0.00±0.00 \\
    \multicolumn{1}{c|}{} & \multicolumn{1}{c|}{\multirow{2}[0]{*}{NPP}} & ACC@1↑ & 0.03±0.00 & 0.05±0.01 & 0.03±0.00 & 0.05±0.02 & 0.08±0.01 & \textbf{0.11±0.04} & \underline{0.10±0.03} \\
    \multicolumn{1}{c|}{} & \multicolumn{1}{c|}{} & ACC@5↑ & 0.06±0.01 & 0.10±0.00 & 0.07±0.01 & 0.11±0.03 & \underline{0.18±0.01} & \textbf{0.18±0.05} & 0.14±0.05 \\
    \multicolumn{1}{c|}{} & \multicolumn{1}{c|}{\multirow{2}[1]{*}{TUL}} & ACC@1↑ & 0.26±0.06 & 0.31±0.01 & 0.27±0.06 & 0.27±0.08 & 0.40±0.01 & \underline{0.40±0.05} & \textbf{0.44±0.03} \\
    \multicolumn{1}{c|}{} & \multicolumn{1}{c|}{} & F1↑   & 0.11±0.04 & 0.17±0.01 & 0.14±0.01 & 0.16±0.01 & 0.18±0.01 & \underline{0.18±0.07} & \textbf{0.25±0.00} \\
    \midrule
    \multicolumn{1}{c|}{\multirow{6}[2]{*}{\rotatebox{90}{San Francisco}}} & \multicolumn{1}{c|}{\multirow{2}[1]{*}{POIC}} & ACC@1↑ & 0.06±0.03 & \underline{0.01±0.00} & 0.06±0.01 & \textbf{0.07±0.02} & 0.04±0.01 & 0.06±0.02 & 0.06±0.01 \\
    \multicolumn{1}{c|}{} & \multicolumn{1}{c|}{} & F1↑   & 0.00±0.00 & 0.01±0.01 & 0.00±0.00 & 0.00±0.00 & \textbf{0.02±0.01} & \underline{0.01±0.01} & 0.00±0.00 \\
    \multicolumn{1}{c|}{} & \multicolumn{1}{c|}{\multirow{2}[0]{*}{NPP}} & ACC@1↑ & 0.02±0.01 & 0.01±0.00 & 0.02±0.00 & \underline{0.04±0.01} & 0.03±0.01 & 0.03±0.03 & \textbf{0.06±0.00} \\
    \multicolumn{1}{c|}{} & \multicolumn{1}{c|}{} & ACC@5↑ & 0.06±0.01 & 0.06±0.01 & 0.06±0.01 & \underline{0.10±0.01} & 0.06±0.01 & 0.06±0.06 & \textbf{0.12±0.01} \\
    \multicolumn{1}{c|}{} & \multicolumn{1}{c|}{\multirow{2}[1]{*}{TUL}} & ACC@1↑ & 0.25±0.08 & 0.34±0.01 & 0.30±0.11 & 0.35±0.12 & 0.52±0.01 & \underline{0.53±0.06} & \textbf{0.61±0.01} \\
    \multicolumn{1}{c|}{} & \multicolumn{1}{c|}{} & F1↑   & 0.13±0.01 & 0.15±0.00 & 0.10±0.06 & 0.15±0.05 & 0.26±0.01 & \underline{0.29±0.02} & \textbf{0.32±0.01} \\
    \midrule
    \multicolumn{3}{c|}{Per. Avg. Rank} & 5.8   & 5.7   & 4.5   & 4.2   & 3.0   & \underline{2.9} & \textbf{2.4} \\
    \bottomrule
    \end{tabular}%
    }
  \label{tab:poi}%
  \vspace{-4mm}
\end{table}%

\subsection{Overall Experimental Results}~\label{sec:42}

\begin{table}[t]
    \centering
    \vspace{-0.3cm}
    \caption{\updatetw{Performance comparison of Road Segment MapRL methods on various datasets.}}
    \vspace{-0.3cm}
    \setlength{\tabcolsep}{0.4mm}
    \resizebox{\linewidth}{!}{ 
    \begin{tabular}{ccc|ccccccc}
    \toprule
    \multicolumn{1}{c|}{Data} & \multicolumn{1}{c|}{Task} & Metric & RN2Vec & HRNR  & SARN  & Toast & HRoad & START & JCLRNT \\
    \midrule
    \multicolumn{1}{c|}{\multirow{6}[2]{*}{\rotatebox{90}{Beijing}}} & \multicolumn{1}{c|}{\multirow{2}[1]{*}{ASI}} & MAE↓  & 3.58±0.01 & 3.16±0.00 & 2.81±0.00 & 2.74±0.01 & \textbf{2.71±0.01} & 2.72±0.01 & \underline{2.80±0.02} \\
    \multicolumn{1}{c|}{} & \multicolumn{1}{c|}{} & RMSE↓ & 5.73±0.00 & 5.59±0.00 & 5.48±0.00 & 5.51±0.01 & 5.41±0.00 & \textbf{5.39±0.00} & \underline{5.48±0.01} \\
    \multicolumn{1}{c|}{} & \multicolumn{1}{c|}{\multirow{2}[0]{*}{TTE}} & MAE↓  & 661±14.0 & 631±19.3 & 596±20.8 & \underline{533±33.0} & 585±16.3 & \textbf{512±19.9} & 514±38.3 \\
    \multicolumn{1}{c|}{} & \multicolumn{1}{c|}{} & RMSE↓ & 2978±48 & 2968±42 & 2581±84 & 2576±79 & 2554±35 & \underline{2502±37} & \textbf{2464±36} \\
    \multicolumn{1}{c|}{} & \multicolumn{1}{c|}{\multirow{2}[1]{*}{STS}} & ACC@3↑ & 0.76±0.03 & 0.82±0.03 & 0.84±0.03 & 0.79±0.10 & 0.87±0.01 & \textbf{0.90±0.03} & \underline{0.90±0.04} \\
    \multicolumn{1}{c|}{} & \multicolumn{1}{c|}{} & MR↓   & 12.4±1.54 & \underline{11.5±1.94} & 13.30±2.17 & 12.8±2.81 & 14.0±1.42 & 10.9±1.98 & \textbf{10.4±1.15} \\
    \midrule
    \multicolumn{1}{c|}{\multirow{6}[2]{*}{\rotatebox{90}{Chendu}}} & \multicolumn{1}{c|}{\multirow{2}[1]{*}{ASI}} & MAE↓  & 6.57±0.06 & 6.42±0.04 & 5.99±0.02 & 6.36±0.03 & \textbf{5.93±0.02} & 6.05±0.01 & \underline{6.35±0.08} \\
    \multicolumn{1}{c|}{} & \multicolumn{1}{c|}{} & RMSE↓ & 14.0±0.07 & 13.7±0.07 & 13.3±0.04 & 13.3±0.16 & \textbf{13.2±0.00} & \underline{13.5±0.01} & 13.5±0.13 \\
    \multicolumn{1}{c|}{} & \multicolumn{1}{c|}{\multirow{2}[0]{*}{TTE}} & MAE↓  & 106±0.06 & \underline{100±0.04} & \textbf{88.3±4.27} & 95.0±2.44 & 88.5±5.48 & 95.8±3.12 & 97.0±4.08 \\
    \multicolumn{1}{c|}{} & \multicolumn{1}{c|}{} & RMSE↓ & 155±0.07 & 150±0.05 & 157±6.70 & 150±2.75 & \textbf{137±5.67} & \underline{151±3.84} & 148±3.79 \\
    \multicolumn{1}{c|}{} & \multicolumn{1}{c|}{\multirow{2}[1]{*}{STS}} & ACC@3↑ & 0.69±0.04 & \underline{0.73±0.06} & 0.68±0.13 & 0.69±0.02 & 0.68±0.05 & \textbf{0.76±0.03} & 0.72±0.00 \\
    \multicolumn{1}{c|}{} & \multicolumn{1}{c|}{} & MR↓   & 48.1±1.67 & 43.0±3.37 & 36.6±3.14 & 22.4±3.66 & 24.4±3.01 & \underline{19.2±2.72} & \textbf{18.7±2.53} \\
    \midrule
    \multicolumn{1}{c|}{\multirow{6}[2]{*}{\rotatebox{90}{Xi'an}}} & \multicolumn{1}{c|}{\multirow{2}[1]{*}{ASI}} & MAE↓  & 6.33±0.03 & \underline{5.79±0.12} & \textbf{5.36±0.03} & 5.63±0.01 & 5.41±0.07 & 5.38±0.07 & 5.67±0.02 \\
    \multicolumn{1}{c|}{} & \multicolumn{1}{c|}{} & RMSE↓ & 13.1±0.37 & \underline{11.2±0.10} & \textbf{10.6±0.01} & 10.8±0.12 & 10.7±0.05 & 10.8±0.07 & 10.9±0.06 \\
    \multicolumn{1}{c|}{} & \multicolumn{1}{c|}{\multirow{2}[0]{*}{TTE}} & MAE↓  & 137±6.67 & 134±5.09 & \underline{110±4.26} & 122±4.52 & 110±3.68 & 115±15.4 & \textbf{108±6.82} \\
    \multicolumn{1}{c|}{} & \multicolumn{1}{c|}{} & RMSE↓ & \underline{199±8.23} & 203±11.9 & \textbf{169±5.63} & 191±3.62 & 172±8.94 & 170±13.9 & 178±8.34 \\
    \multicolumn{1}{c|}{} & \multicolumn{1}{c|}{\multirow{2}[1]{*}{STS}} & ACC@3↑ & 0.71±0.01 & 0.74±0.04 & 0.72±0.07 & \textbf{0.80±0.03} & 0.73±0.01 & 0.75±0.04 & \underline{0.77±0.02} \\
    \multicolumn{1}{c|}{} & \multicolumn{1}{c|}{} & MR↓   & 12.9±1.71 & \underline{12.1±1.48} & 12.5±3.62 & 16.5±4.28 & 19.0±2.95 & \textbf{10.8±3.54} & 11.9±3.15 \\
    \midrule
    \multicolumn{1}{c|}{\multirow{6}[2]{*}{\rotatebox{90}{Porto}}} & \multicolumn{1}{c|}{\multirow{2}[1]{*}{ASI}} & MAE↓  & 4.39±0.00 & 4.40±0.00 & 4.29±0.00 & 4.36±0.01 & \textbf{4.20±0.01} & 4.21±0.00 & \underline{4.35±0.05} \\
    \multicolumn{1}{c|}{} & \multicolumn{1}{c|}{} & RMSE↓ & 7.90±0.00 & \textbf{7.58±0.00} & 7.78±0.00 & \underline{7.93±0.05} & 7.71±0.01 & 7.79±0.04 & 7.79±0.05 \\
    \multicolumn{1}{c|}{} & \multicolumn{1}{c|}{\multirow{2}[0]{*}{TTE}} & MAE↓  & 99.5±1.78 & 101±1.98 & \underline{97.2±1.38} & 98.4±1.81 & 98.5±2.83 & 97.5±6.92 & \textbf{96.8±1.30} \\
    \multicolumn{1}{c|}{} & \multicolumn{1}{c|}{} & RMSE↓ & 150±3.60 & 147±2.60 & \underline{143±1.80} & 147±2.48 & \textbf{140±2.22} & 158±7.04 & 145±1.03 \\
    \multicolumn{1}{c|}{} & \multicolumn{1}{c|}{\multirow{2}[1]{*}{STS}} & ACC@3↑ & 0.80±0.10 & 0.82±0.11 & 0.83±0.03 & \textbf{0.90±0.06} & 0.83±0.04 & \underline{0.88±0.02} & 0.87±0.04 \\
    \multicolumn{1}{c|}{} & \multicolumn{1}{c|}{} & MR↓   & 14.3±1.91 & \underline{12.0±1.81} & 17.9±3.40 & 12.1±4.03 & 17.2±1.80 & 13.7±2.44 & \textbf{11.4+3.38} \\
    \midrule
    \multicolumn{1}{c|}{\multirow{6}[2]{*}{\rotatebox{90}{San Francisco}}} & \multicolumn{1}{c|}{\multirow{2}[1]{*}{ASI}} & MAE↓  & 2.98±0.00 & 2.51±0.00 & 2.49±0.01 & 2.56±0.00 & 2.48±0.01 & \underline{2.45±0.00} & \textbf{2.42±0.01} \\
    \multicolumn{1}{c|}{} & \multicolumn{1}{c|}{} & RMSE↓ & 5.80±0.00 & 5.38±0.01 & \underline{5.29±0.01} & 5.29±0.00 & \textbf{5.25±0.00} & 5.33±0.00 & 5.37±0.00 \\
    \multicolumn{1}{c|}{} & \multicolumn{1}{c|}{\multirow{2}[0]{*}{TTE}} & MAE↓  & 289±1.02 & 288±1.01 & 279±3.21 & 274±9.01 & \underline{270±7.61} & \textbf{268±10.2} & 275±9.30 \\
    \multicolumn{1}{c|}{} & \multicolumn{1}{c|}{} & RMSE↓ & 1479±4.3 & 1449±4.4 & 1499±3.6 & \textbf{1372±28} & 1451±16 & 1497±38 & \underline{1446±21} \\
    \multicolumn{1}{c|}{} & \multicolumn{1}{c|}{\multirow{2}[1]{*}{STS}} & ACC@3↑ & 0.84±0.04 & 0.91±0.00 & 0.86±0.07 & \textbf{0.94±0.04} & \underline{0.92±0.01} & 0.90±0.07 & 0.87±0.05 \\
    \multicolumn{1}{c|}{} & \multicolumn{1}{c|}{} & MR↓   & 7.20±3.19 & 7.00±0.01 & 8.48±2.05 & 9.22±3.29 & \underline{6.55±1.22} & 6.71±2.01 & \textbf{6.49±1.63} \\
    \midrule
    \multicolumn{3}{c|}{Per Avg Rank} & 6.3   & 4.9   & 3.9   & 3.9   & 3.2   & \underline{2.9} & \textbf{2.8} \\
    \bottomrule
    \end{tabular}%
}
\label{tab:road}
\vspace{-0.3cm}
\end{table}

\subsubsection{Experimental Results for POI \rl} Table~\ref{tab:poi} shows the performance of \rl models on three downstream tasks: POI Classification (POIC), Next POI Prediction (NPP), and Trajectory User Link (TUL), each evaluated using two types of metrics. The table also reports average performance rankings across tasks (\underline{Pre Ave Rank}) and the ranking based on the models' average performance (\underline{Pre Rank}). From Tab.~\ref{tab:poi}, we have the following observations.

1) {\em Importance of Sequence-based Encoders}: Models integrating token- and sequence-based encoders (\eg, CTLE, CACSR, Hier) outperform token-based models, indicating the value of trajectory data. Among these, models using Augmented Trajectory Contrastive Learning (ATrCL) and Masked Trajectory Recovery (MTR) tasks (\eg, CTLE, CACSR) achieve better performance than those focused on Trajectory Prediction (TrajP) tasks (\eg, Hier). This is likely because ATrCL and MTR capture long-range temporal dependencies across entire trajectories, while TrajP focuses only on consecutive POI dependencies.

2) {\em Impact of Relation Networks}: 
For token-based models, those incorporating relation networks (\eg, Teaser, P2Vec) outperform models without them (\eg, SkipGram, Tale), highlighting their value. However, for models combining token- and sequence-based encoders, relation networks provide little benefit. This is likely due to the sparsity of POI trajectories, where the number of trajectories between two POIs is often dictated by geographical proximity. As a result, the information provided by relation networks overlaps significantly with that from trajectory data. Once the dependency within the trajectory data has been fully exploited, the relational network contributes limited additional information.\label{sec:t42}

3) {\em TokRI vs. TRCL Tasks}: Token-based models utilizing the Token Relation Inference (TokRI) task, such as P2Vec, generally outperform those employing Token Relation Contrastive Learning (TRCL), such as SkipGram and Teaser. With sufficient data, supervised learning via TokRI effectively leverages labeled data to generate more discriminative representations. In contrast, the indirect nature of contrastive learning in TRCL offers less direct utilization of labeled information, resulting in comparatively lower model performance.

\updatetw{
4) {\em Instability Introduced by Contrastive Learning.} Models that rely on contrastive learning tasks, such as SkipGram, Teaser, and CACSR, exhibit instability due to their sensitivity to dataset partitioning and negative sample selection. In contrast, P2Vec and Tale, which use TokRI—a supervised pretraining task—demonstrate greater stability by explicitly modeling relationships between map entities.}



\begin{table}[t]
  \centering
  \vspace{-0.3cm}
  \caption{\updatetw{Performance comparison of land parcel \rl methods on various datasets.}}
  \vspace{-0.3cm}
  \setlength{\tabcolsep}{0.4mm}
    \resizebox{0.97\linewidth}{!}{ 
    \begin{tabular}{ccc|ccccccc}
    \toprule
    \multicolumn{1}{c|}{Data} & \multicolumn{1}{c|}{Task} & Metric & ZEMob & GMEL  & HDGE  & MVURE & MGFN  & ReMVC & HREP \\
    \midrule
    \multicolumn{1}{c|}{\multirow{6}[2]{*}{\rotatebox{90}{Beijing}}} & \multicolumn{1}{c|}{\multirow{2}[1]{*}{LPC}} & ACC@1↑ & 0.43±0.03 & 0.46±0.03 & 0.41±0.00 & 0.48±0.00 & \underline{0.59±0.02} & \textbf{0.75±0.02} & 0.50±0.01 \\
    \multicolumn{1}{c|}{} & \multicolumn{1}{c|}{} & F1↑   & 0.11±0.00 & 0.61±0.03 & 0.22±0.01 & 0.28±0.01 & \underline{0.65±0.04} & \textbf{0.71±0.02} & 0.38±0.01 \\
    \multicolumn{1}{c|}{} & \multicolumn{1}{c|}{\multirow{2}[0]{*}{FI}} & MAE↓  & 30.9±0.03 & 27.7±0.01 & \underline{20.8±2.91} & 29.6±0.83 & 30.1±0.02 & 27.8±0.02 & \textbf{20.6±0.33} \\
    \multicolumn{1}{c|}{} & \multicolumn{1}{c|}{} & RMSE↓ & 55.3±0.04 & 59.0±0.04 & \underline{44.8±2.79} & 53.7±1.36 & 54.5±0.02 & 52.7±0.04 & \textbf{42.6±0.86} \\
    \multicolumn{1}{c|}{} & \multicolumn{1}{c|}{\multirow{2}[1]{*}{MI}} & MAE↓  & 0.43±0.03 & 0.46±0.03 & 0.34±0.03 & \textbf{0.27±0.01} & 0.30±0.03 & \underline{0.30±0.01} & 0.32±0.00 \\
    \multicolumn{1}{c|}{} & \multicolumn{1}{c|}{} & RMSE↓ & 0.80±0.02 & 0.74±0.02 & 0.70±0.03 & 0.71±0.02 & 0.76±0.04 & \underline{0.69±0.02} & \textbf{0.66±0.01} \\
    \midrule
    \multicolumn{1}{c|}{\multirow{6}[2]{*}{\rotatebox{90}{Chengdu}}} & \multicolumn{1}{c|}{\multirow{2}[1]{*}{LPC}} & ACC@1↑ & 0.36±0.01 & 0.38±0.04 & 0.42±0.03 & 0.48±0.00 & 0.48±0.08 & \textbf{0.66±0.03} & \underline{0.54±0.01} \\
    \multicolumn{1}{c|}{} & \multicolumn{1}{c|}{} & F1↑   & 0.17±0.03 & 0.16±0.03 & 0.27±0.06 & 0.27±0.01 & 0.35±0.12 & \textbf{0.59±0.03} & \underline{0.46±0.01} \\
    \multicolumn{1}{c|}{} & \multicolumn{1}{c|}{\multirow{2}[0]{*}{FI}} & MAE↓  & 155±0.03 & \underline{155±0.02} & 159±2.02 & 156±3.10 & 158±5.69 & \textbf{148±0.02} & 156±2.8 \\
    \multicolumn{1}{c|}{} & \multicolumn{1}{c|}{} & RMSE↓ & 441±0.02 & 441±0.04 & 403±1.12 & 443±4.28 & \textbf{337±24.4} & 431±0.03 & \underline{396±9.7} \\
    \multicolumn{1}{c|}{} & \multicolumn{1}{c|}{\multirow{2}[1]{*}{MI}} & MAE↓  & \textbf{3.64±0.03} & 4.62±0.04 & 5.17±0.01 & \underline{3.97±0.71} & 5.14±0.02 & 5.29±0.04 & 5.47±0.12 \\
    \multicolumn{1}{c|}{} & \multicolumn{1}{c|}{} & RMSE↓ & 29.7±0.02 & 29.7±0.03 & \underline{28.7±0.15} & 29.6±0.12 & 29.3±1.14 & 29.1±0.06 & \textbf{28.3±0.07} \\
    \midrule
    \multicolumn{1}{c|}{\multirow{6}[2]{*}{\rotatebox{90}{Xi'an}}} & \multicolumn{1}{c|}{\multirow{2}[1]{*}{LPC}} & ACC@1↑ & 0.36±0.03 & 0.36±0.03 & 0.44±0.02 & 0.53±0.01 & 0.48±0.09 & \underline{0.57±0.00} & \textbf{0.59±0.01} \\
    \multicolumn{1}{c|}{} & \multicolumn{1}{c|}{} & F1↑   & 0.19±0.04 & 0.42±0.03 & 0.25±0.01 & 0.26±0.02 & 0.28±0.07 & \textbf{0.46±0.03} & \underline{0.45±0.02} \\
    \multicolumn{1}{c|}{} & \multicolumn{1}{c|}{\multirow{2}[0]{*}{FI}} & MAE↓  & 134±0.03 & 133±0.02 & \underline{124±5.87} & 127±3.85 & \textbf{119±9.30} & 132±0.04 & 129±0.80 \\
    \multicolumn{1}{c|}{} & \multicolumn{1}{c|}{} & RMSE↓ & 326±0.02 & 324±0.03 & \underline{289±26.3} & 318±4.77 & \textbf{241±28.4} & 325±0.03 & 297±9.10 \\
    \multicolumn{1}{c|}{} & \multicolumn{1}{c|}{\multirow{2}[1]{*}{MI}} & MAE↓  & 3.83±0.03 & 3.71±0.04 & 3.92±0.03 & \textbf{3.33±0.35} & \underline{3.70±0.11} & 3.91±0.05 & 4.02±0.04 \\
    \multicolumn{1}{c|}{} & \multicolumn{1}{c|}{} & RMSE↓ & 17.7±0.03 & 17.6±0.03 & \textbf{16.6±0.14} & 17.4±0.00 & 16.8±1.32 & 16.8±0.05 & \underline{16.7±0.20} \\
    \midrule
    \multicolumn{1}{c|}{\multirow{6}[2]{*}{\rotatebox{90}{Porto}}} & \multicolumn{1}{c|}{\multirow{2}[1]{*}{LPC}} & ACC@1↑ & 0.34±0.04 & 0.33±0.04 & \underline{0.42±0.05} & 0.30±0.01 & 0.30±0.01 & \textbf{0.44±0.03} & 0.36±0.02 \\
    \multicolumn{1}{c|}{} & \multicolumn{1}{c|}{} & F1↑   & 0.21±0.01 & 0.38±0.03 & \underline{0.41±0.07} & 0.27±0.02 & 0.20±0.03 & \textbf{0.41±0.04} & 0.32±0.03 \\
    \multicolumn{1}{c|}{} & \multicolumn{1}{c|}{\multirow{2}[0]{*}{FI}} & MAE↓  & 528±0.03 & 525±0.01 & 501±18.0 & \underline{402±21.8} & 477±37.3 & 527±0.01 & \textbf{289±12.1} \\
    \multicolumn{1}{c|}{} & \multicolumn{1}{c|}{} & RMSE↓ & 595±0.02 & 592±0.04 & 582±14.4 & 472±21.1 & \underline{443±36.8} & 595±0.03 & \textbf{355±27.3} \\
    \multicolumn{1}{c|}{} & \multicolumn{1}{c|}{\multirow{2}[1]{*}{MI}} & MAE↓  & 44.9±0.02 & 41.1±0.03 & 30.8±4.57 & 37.8±0.65 & \underline{29.8±7.26} & 39.4±0.05 & \textbf{28.3±5.30} \\
    \multicolumn{1}{c|}{} & \multicolumn{1}{c|}{} & RMSE↓ & 72.3±0.02 & 56.2±0.04 & 45.7±4.91 & 54.1±4.68 & \underline{45.5±9.79} & 56.7±0.03 & \textbf{42.2±8.93} \\
    \midrule
    \multicolumn{1}{c|}{\multirow{6}[2]{*}{\rotatebox{90}{San Francisco}}} & \multicolumn{1}{c|}{\multirow{2}[1]{*}{LPC}} & ACC@1↑ & \underline{0.84±0.02} & 0.64±0.01 & 0.82±0.01 & 0.84±0.02 & 0.83±0.01 & \textbf{0.86±0.03} & 0.84±0.01 \\
    \multicolumn{1}{c|}{} & \multicolumn{1}{c|}{} & F1↑   & 0.53±0.03 & \underline{0.56±0.01} & 0.50±0.03 & 0.52±0.05 & 0.44±0.10 & 0.47±0.04 & \textbf{0.57±0.03} \\
    \multicolumn{1}{c|}{} & \multicolumn{1}{c|}{\multirow{2}[0]{*}{FI}} & MAE↓  & 509±0.04 & 514±0.01 & 387±23.1 & \underline{349±23.7} & 356±46 & 445±0.01 & \textbf{324±25.8} \\
    \multicolumn{1}{c|}{} & \multicolumn{1}{c|}{} & RMSE↓ & 910±0.04 & 917±0.02 & \underline{693±35.8} & 700±44.2 & 748±80 & 852±0.02 & \textbf{624±31} \\
    \multicolumn{1}{c|}{} & \multicolumn{1}{c|}{\multirow{2}[1]{*}{MI}} & MAE↓  & \textbf{5.88±0.03} & 7.40±0.02 & 6.51±0.51 & 6.46±0.42 & 6.40±0.84 & 6.87±0.04 & \underline{5.89±0.50} \\
    \multicolumn{1}{c|}{} & \multicolumn{1}{c|}{} & RMSE↓ & 23.5±0.01 & 22.3±0.01 & 21.5±0.28 & 20.6±1.21 & \underline{20.2±2.34} & 22.7±0.03 & \textbf{19.1±1.51} \\
    \midrule
    \multicolumn{3}{c|}{Per Avg Rank} & 5.67  & 5.10  & 4.00  & 3.73  & 3.53  & \underline{3.50} & \textbf{2.43} \\
    \bottomrule
    \end{tabular}%
    }
  \label{tab:region}%
  \vspace{-0.4cm}
\end{table}%

\subsubsection{Experimental Results for Road Segment \rl}

Table~\ref{tab:road} shows a performance comparison of road segment \rl models across three downstream tasks: Average Speed Inference (ASI), Travel Time Estimation (TTE), and Similarity Trajectory Search (STS). The following key observations can be drawn from the results:

1) {\em Full Pipeline Models Perform Best}: Models utilizing the complete token-graph-sequence pipeline, such as JCLRNT and START, achieve the highest performance compared to models with partial pipelines. Additionally, token-based + graph-based and token-based + sequence-based models -- such as HRNR, SARN, Toast, and HRoad -- outperform purely token-based models like RN2Vec, highlighting the value of integrating trajectory and relation network information. Graph-based and sequence-based encoders capture essential patterns that improve road segment representations.

2) {\em Multiple Pre-training Tasks Improve Performance}: Models incorporating multiple pre-training tasks, such as Toast, HRoad, START, and JCLRNT, consistently outperform those using a single pre-training task, like HRNR and SARN. Among token-based + graph-based models, HRoad, which integrates both NFI and GAu tasks, surpasses HRNR (GAu only) and SARN (AGCL only), highlighting the benefits of leveraging diverse pre-training tasks. By extracting information from multiple perspectives, these models learn more comprehensive feature representations. Additionally, in full-pipeline models, JCLRNT, which combines a sequence-based task (Augmented Trajectory Contrastive Learning) with a graph-based task (Augmented Graph Contrastive Learning), outperforms START, which employs two sequence-based tasks (MTR and ATrCL). This suggests that even among multi-task models, greater task diversity further enhances performance.

\updatetw{
3) {\em Fluctuating Results for SARN, Toast, START, and JCLRNT:}
These models experience higher performance variability compared to other models, largely due to their dependence on contrastive pretraining techniques such as Token Relation Contrastive Learning (TRCL), Augmented Graph Contrastive Learning (AGCL), and Augmented Trajectory Contrastive Learning (ATrCL). These methods are particularly vulnerable to shifts in training data splits and the inherent randomness in negative sample selection, leading to inconsistent results across multiple runs.
}


\subsubsection{Experimental Results for Land Parcel \rl}

Table~\ref{tab:region} presents the performance of land parcel \rl models on the downstream tasks of Land Parcel Classification (LPC), Flow Inference (FI), and Mobility Inference (MI). The following observations can be made:

1) {\em Incorporating POI Data Enhances Performance}: Models that utilize POI data, such as MVURE, ReMVC, and HREP (see the ``Map Entities'' column in Table~\ref{tab:models}), outperform those that do not. These models treat POI categories within a land parcel as features, enriching the representation of the parcel's functional characteristics. This approach provides valuable contextual information, improving the overall performance of the models.

2) {\em Limitations of Current Encoder Structures}: Most existing land parcel \rl models adopt a token-based + graph-based encoder structure, focusing on extracting information from relation networks. However, these models overlook the sequential dependencies within trajectories, limiting their effectiveness. Future research should explore advanced methods to capture trajectory sequence information relevant to land parcels for further improvements.


\updatetw{
3) {\em Fluctuating Results for MGFN}: 
Although MGFN ranks third overall (with an average rank of 3.53), its performance varies considerably under different random seeds, indicating a degree of instability. This instability arises because MGFN relies on clustering techniques to build the base relation network for representation learning, making it more sensitive to randomness.

4) {\em Small Datasets Mitigate Instability in Negative Sampling}: ZEMob, GMEL, and ReMVC are among the most stable models across all datasets. Despite employing contrastive learning as pretraining tasks—which tends to be sensitive in other MapRL models—their stability may be attributed to the smaller number of parcel entities. For instance, San Francisco contains 194 parcel entities compared to 27,274 road segment entities. This smaller number of parcel entities likely results in more consistent negative sampling. Additionally, tasks for land parcels are simpler, focusing on static attribute inference rather than complex sequential predictions, which further contributes to stable learning.
}

\subsubsection{Summary for Overall Experiments} The overall experimental results highlight the following key findings across POI, road segment, and land parcel \rl models.

1) {\em Importance of Combined Encoders}:  Models employing a combination of token-based, graph-based, and sequence-based encoders consistently outperform those relying on single-type encoders. This demonstrates the value of integrating multiple data sources -- such as trajectories and relation networks -- to capture complex spatial, temporal, and relational patterns.

2) {\em Effectiveness of Diverse Pre-training Tasks}: Models that leverage multiple pre-training tasks usually perform better than those limited to a single task. This suggests that diverse pre-training strategies enhance the model's ability to learn comprehensive feature representations from different perspectives.

3) {\em Role of Auxiliary Data}: Incorporating trajectory, and relation network data improves model performance across tasks. However, models that fully exploit trajectory dependencies achieve diminishing gains from relation networks, indicating some redundancy between these data sources.

\updatetw{
4) {Instability Introduced by Contrastive Learning}: Contrastive learning relies on the generation of positive and negative sample pairs, which vary between different random seeds. Consequently, models using contrastive learning pretraining tasks are more sensitive to randomness. However, this sensitivity may decrease when working with smaller datasets.
}

\begin{figure}[t]
    \centering
    \includegraphics[width=0.93\columnwidth]{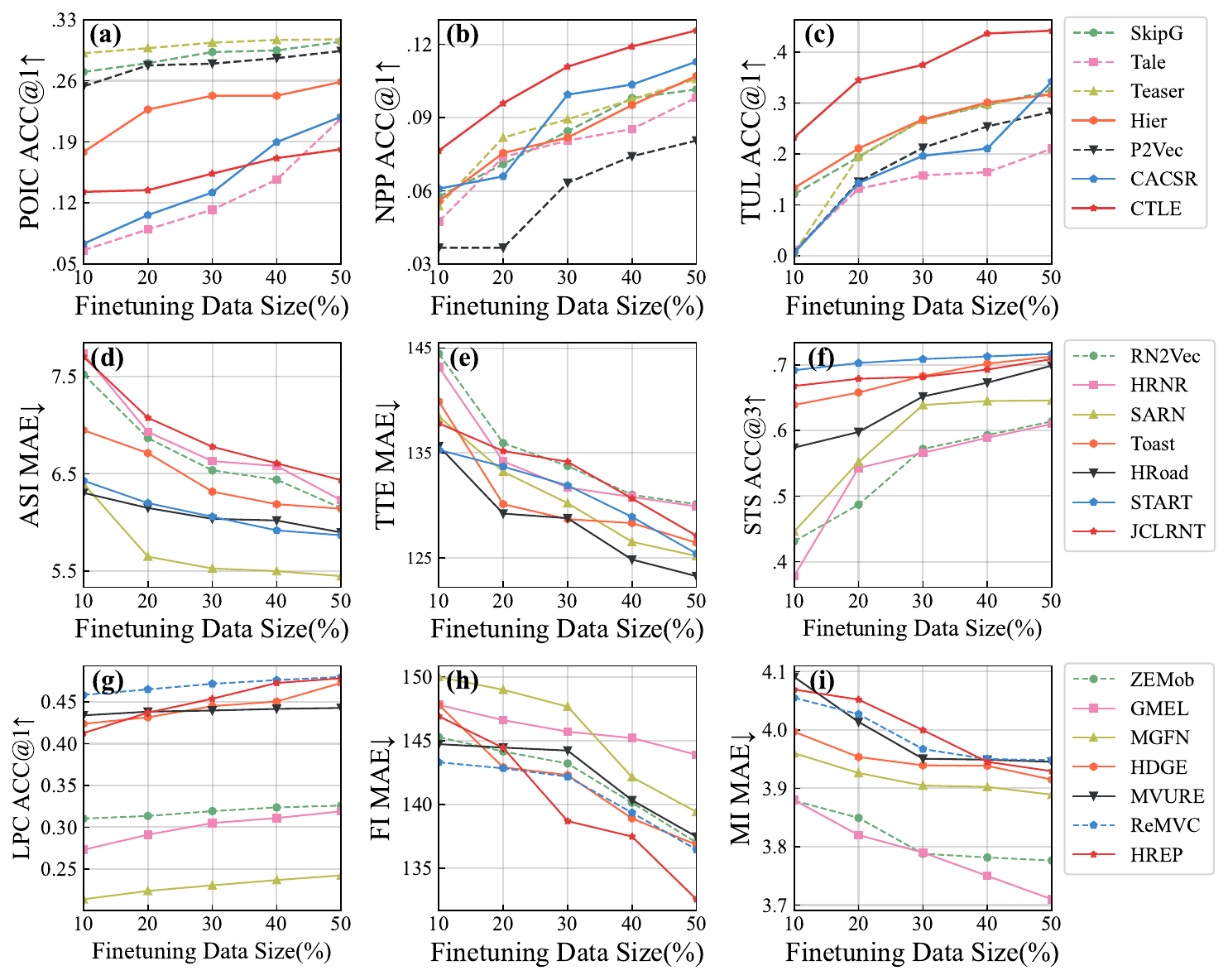}
    \caption{Results from varying fine-tuning data size.}
    \vspace{-0.2cm}
    \label{fig:dive}
    \vspace{-0.4cm}
\end{figure}

\subsection{\updatetr{Scalability Analysis of MapRL Models.}}\label{sec:43}

Scalability is a critical factor in MapRL model design, determining their adaptability to varying data sizes and computational constraints. In this section, we analyze scalability from three key perspectives: effectiveness, efficiency, and the trade-offs between model size and inference time. Effectiveness assesses how well models maintain performance as the amount of fine-tuning data varies. Efficiency evaluates computational costs, including parameter count, training time, and inference time, across datasets with different entity scales. Finally, the trade-off analysis explores how model complexity impacts inference speed, offering insights into balancing accuracy and deployment feasibility. 

\subsubsection{Effectiveness}\label{sec:431}
In this experiment, we adjust the fine-tuning data size from 10\% to 50\% of the original training set and compare the result of pre-built models in \name. POI-oriented models are tested on the Tokyo dataset, while road segment- and land parcel-oriented models are evaluated on the Xi'an dataset. \fig~\ref{fig:dive} presents the evaluation result: \fig~\ref{fig:dive}~(a) to (c) show results for POI-oriented models. \fig~\ref{fig:dive}~(d) to (f) display results for road-segment-oriented models. \fig~\ref{fig:dive}~(g) to (i) provide results for land-parcel-oriented models. From these figures, we make the following observations:\label{sec:t43}

\updatetr{
1) {\em Performance Scalability.}
~Token-based models (dashed lines) show limited scalability, with performance gains plateauing around 30\% of fine-tuning data. In contrast, graph- and sequence-based models continue improving up to 50\%. Notably, methods incorporating contrastive learning techniques—such as CACSR, START, JCLRNT, and HREP—show particularly strong improvements, suggesting that contrastive learning further enhances performance as more data becomes available.
 }

2) {\em Stability in Attribute Inference Tasks}: 
\rl models exhibit stable performance in attribute inference tasks (POIC, ASI, LPC) regardless of training data size. Since these tasks primarily rely on entity attributes learned during pre-training, only a small amount of labeled data is needed for fine-tuning.

\begin{figure}[t]
    \centering
    \includegraphics[width=0.93\linewidth]{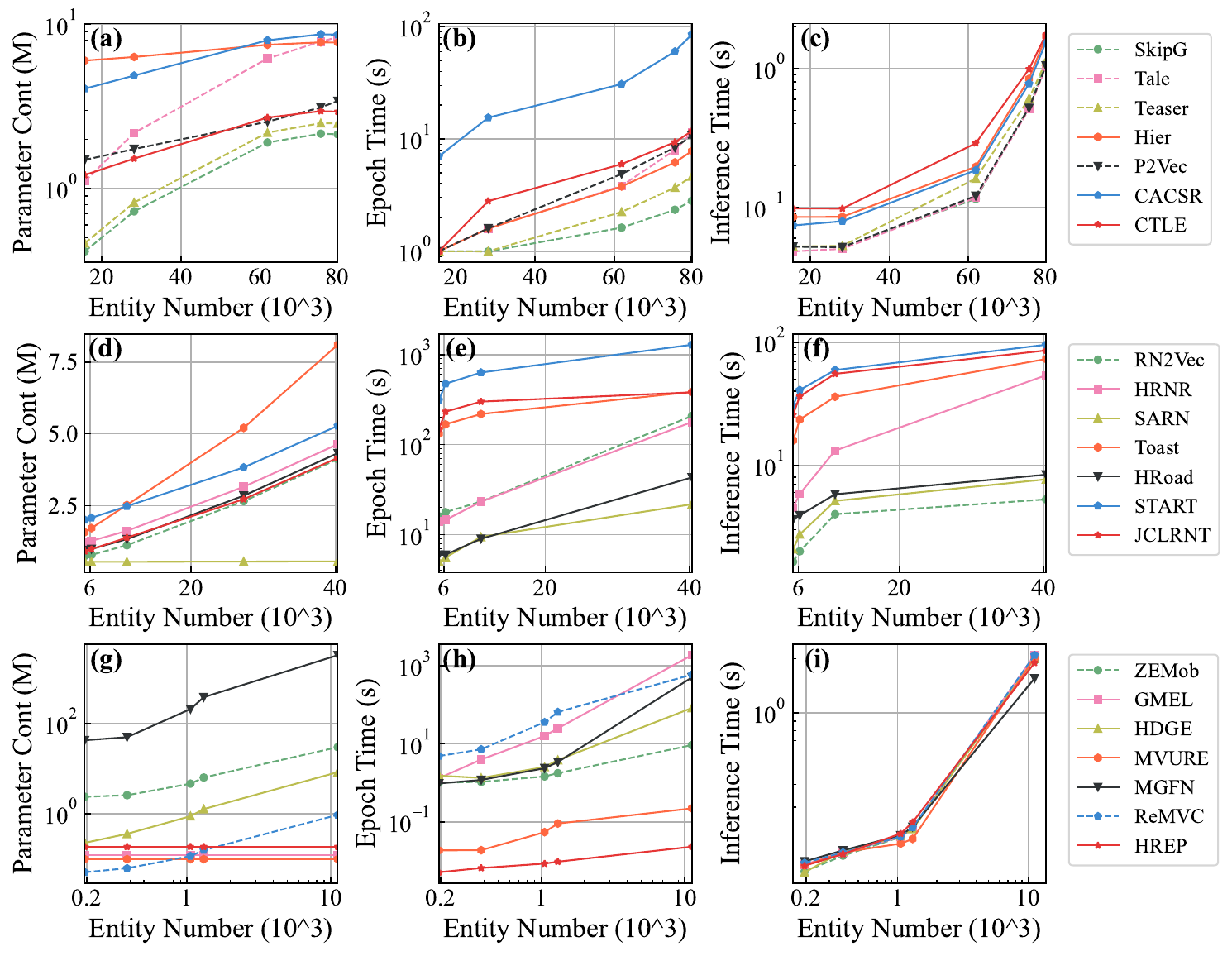}
    \caption{\updatetw{Scalability of MapRL models in model size, training time, and inference time across datasets with varying entity counts. Each point represents a dataset, with lines connecting them in ascending entity count to highlight trends.}}
    \label{fig:efficiency}
    \vspace{-6mm}
\end{figure}

3) {\em Fluctuations in Trajectory-Related and Mobility Inference Tasks}: Models exhibit notable performance fluctuations in trajectory-related tasks (e.g., NPP, TUL, TTE) and mobility inference tasks (e.g., FI, MI). Trajectory-related tasks require sequence-level predictions, whereas most encoders and pre-training tasks generate single-entity embeddings, making these tasks especially sensitive to smaller fine-tuning datasets. By contrast, Similarity Trajectory Search (STS) relies on similarity comparisons rather than prediction, making it more stable. Additionally, Flow Inference (FI) and Mobility Inference (MI) capture human mobility flows that vary greatly over time. This temporal heterogeneity demands more labeled data to accurately model these patterns, leading to further performance variability when data are limited.

\updatetr{
Overall, token-based models show weaker performance scalability, with diminishing returns as dataset size increases. In contrast, graph-based and sequence-based models, especially those utilizing contrastive learning methods (e.g., CACSR, START, JCLRNT, HREP), show greater performance improvements as labeled data increases, demonstrating stronger scalability in data-intensive scenarios. Moreover, existing \rl models exhibit varying sensitivity to the size of labeled data across different tasks. For tasks less influenced by temporal information, such as attribute inference (ASI, POIC, LPC) and Similarity Trajectory Search (STS), models maintain stable performance even with limited training data. However, for tasks with time-dependent labels, such as trajectory-related tasks and mobility inference tasks, larger amounts of labeled data are necessary to effectively capture temporal dynamics.
}

\updatetr{
\subsubsection{Efficiency}\label{sec:432} We evaluate model's efficiency across three key aspects: (1) Parameter count, representing the space complexity of the models; (2) Epoch time, indicating the time cost of training per epoch; and (3) Inference time, reflecting the time required for downstream task evaluation. These measurements are collected from all five datasets used in the main experiments, each with different entity counts. Figure.~\ref{fig:efficiency} presents the evaluation results: \fig~\ref{fig:efficiency} (a) to (c) show results for POI-oriented models, \fig~\ref{fig:efficiency} (d) to (f) display results for road segment-oriented models, and \fig~\ref{fig:efficiency} (g) to (i) provide results for land parcel-oriented models. The horizontal axis of all figures represents the number of entities on a $10^3$ scale, while the vertical axis represents the parameter count in millions, time cost per epoch in seconds, and total inference time in seconds, respectively. Based on these figures, we have several key findings:

 1) {\em Parameter Count:} Token-based models (dashed lines) generally achieve better scalability in model size, while sequence-based models (e.g., CACSR, Toast, START) often have higher parameter counts, likely due to the increased space complexity of sequence encoders. In contrast, SARN, GMEL, MVURE, and HREP show relatively stable parameter counts as the number of entities increases, demonstrating better scalability in scenarios with large numbers of map entities.

\begin{figure}[t]
    \centering
    \includegraphics[width=0.9\linewidth]{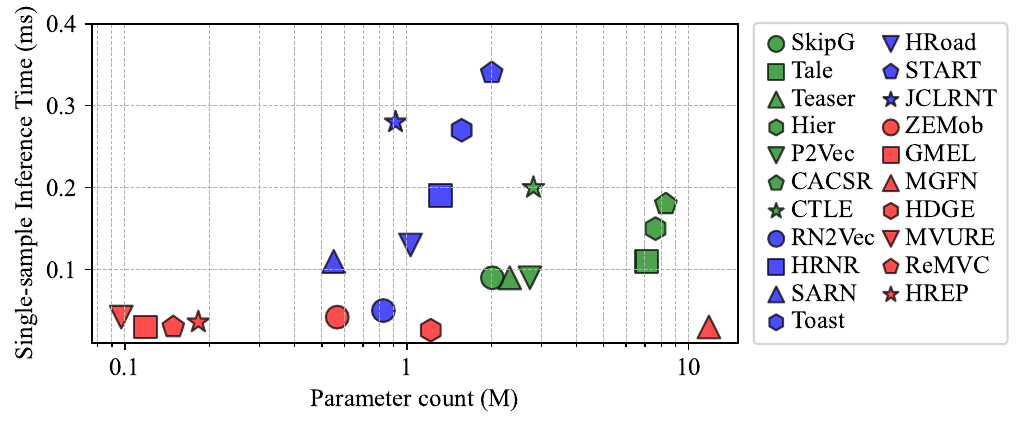}
    \caption{\updatetr{Tradeoff between model parameters and time efficiency on average, across all available datasets.}}
    \label{fig:tradeoff}
    \vspace{-4mm}
\end{figure}

2) {\em Training Time:} Token-based models (dashed lines) demonstrate better training time scalability in POI-oriented models but perform less efficiently in land parcel-oriented models. This difference is influenced by entity count variations, with POI datasets containing $1 \times 10^4$ to $8 \times 10^4$ entities, whereas land parcel datasets range from $2 \times 10^2$ to $1 \times 10^4$. The results suggest that token-based models may have a greater competitive advantage in large-scale datasets. Sequence-based models (e.g., CACSR, Toast, START) generally incur longer training times. Graph-based models employing GAu pretraining tasks (HRNR, GMEL, MGFN) show more-than-linear growth in training time, posing scalability challenges for large datasets, whereas HREP, despite using GAu pretraining, maintains relatively stable training time due to its lower space complexity.

3) {\em Inference Time:} All models show an increasing trend in inference time. However, the rise is more pronounced in POI- and land parcel-oriented models compared to road segment-oriented models, suggesting that inference time is more sensitive to entity count in these tasks. Sequence-based models (e.g., CTLE, Toast, START, JCLRNT) generally require longer inference times. This can be attributed to their reliance on sequential encoders, which introduce additional computational steps.

\subsubsection{Trade-offs.}\label{sec:433}

In Section \ref{sec:experiment}, we observe that complex models tend to achieve higher accuracy (\eg full pipeline models). However, increased model complexity can lead to efficiency challenges, particularly in inference time. The previous efficiency analysis does not directly examine the trade-off between model complexity and inference time. To address this, we evaluate the relationship between parameter count (millions) and single-sample inference time (ms). This experiment uses the same dataset as in Section 4.3.1 on effectiveness: POI-oriented models are evaluated on the Tokyo dataset, while others are tested on the Xi’an dataset.

\fig~\ref{fig:tradeoff} presents the results, showing that land-parcel-oriented models maintain relatively stable inference times despite increases in model size, likely due to the lower complexity of their downstream tasks. In contrast, POI and road segment-oriented models exhibit a clear upward trend in inference time as model size grows. Sequence-based models (e.g., CTLE, JCLRNT, Toast, START) have higher model size and inference times due to the integration of sequence encoders. 
Meanwhile, graph-based models like MVURE, GMEL, ReMVC, and HREP generally maintain lower inference times and smaller model sizes. However, HRNR incurs higher inference costs due to its multi-level GNN encoder.
}

\begin{figure*}
    \centering
    \subfigure[Top 3 combinations for POI map entity.]{\includegraphics[width=0.31\textwidth]{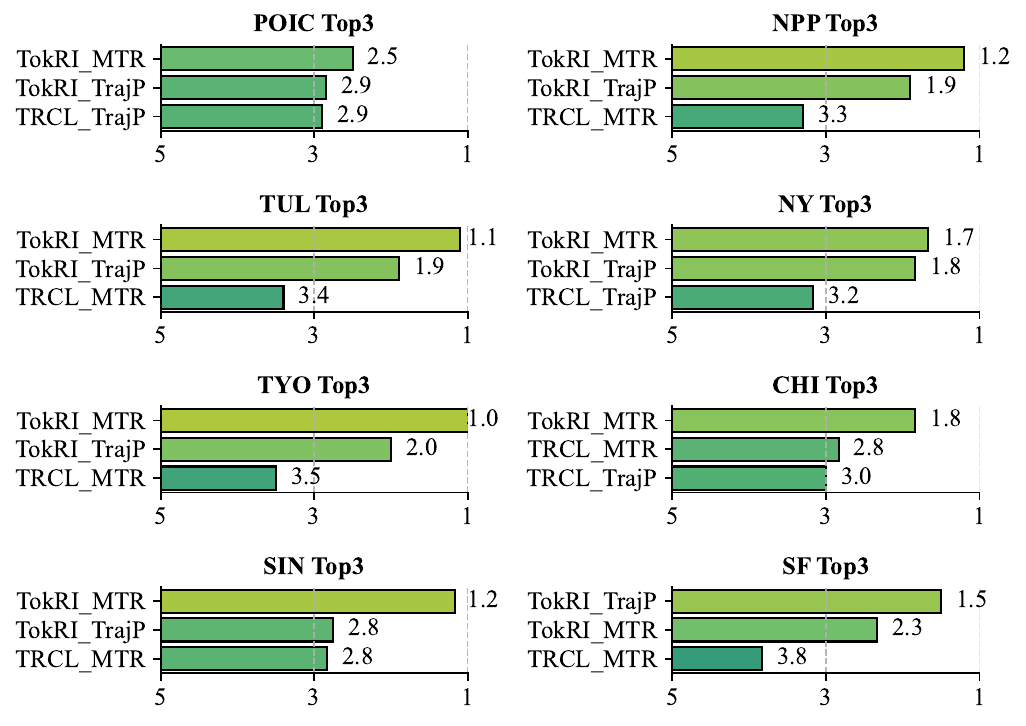}}
    \subfigure[Top 3 combinations for segment map entity.]{\includegraphics[width=0.31\textwidth]{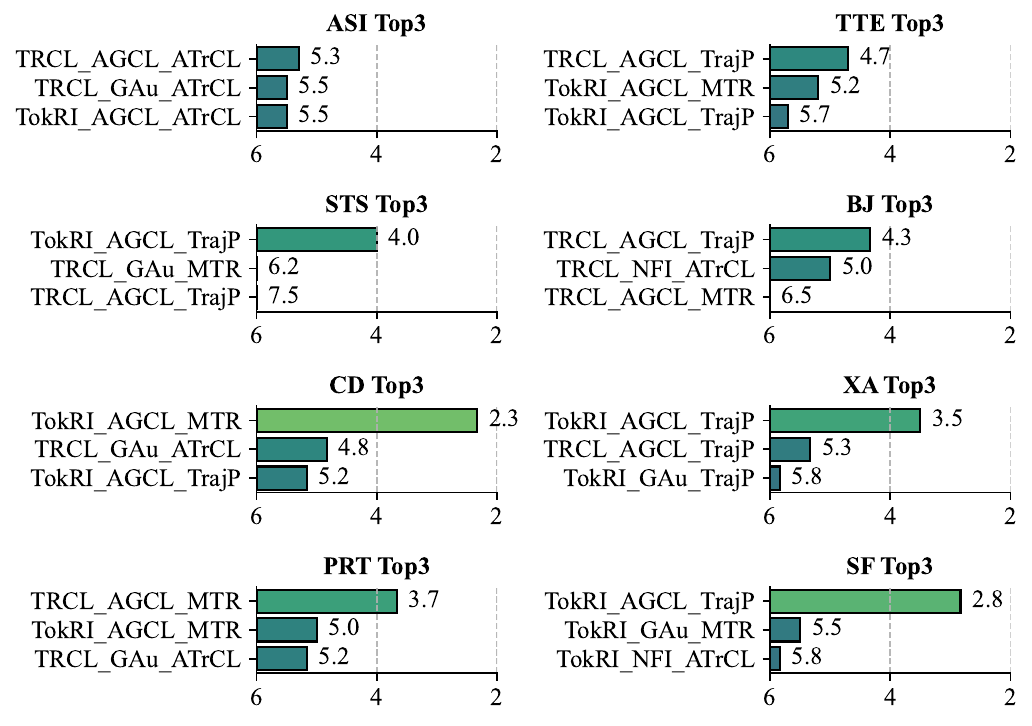}}
    \subfigure[Top 3 combinations for parcel map entity.]{\includegraphics[width=0.31\textwidth]{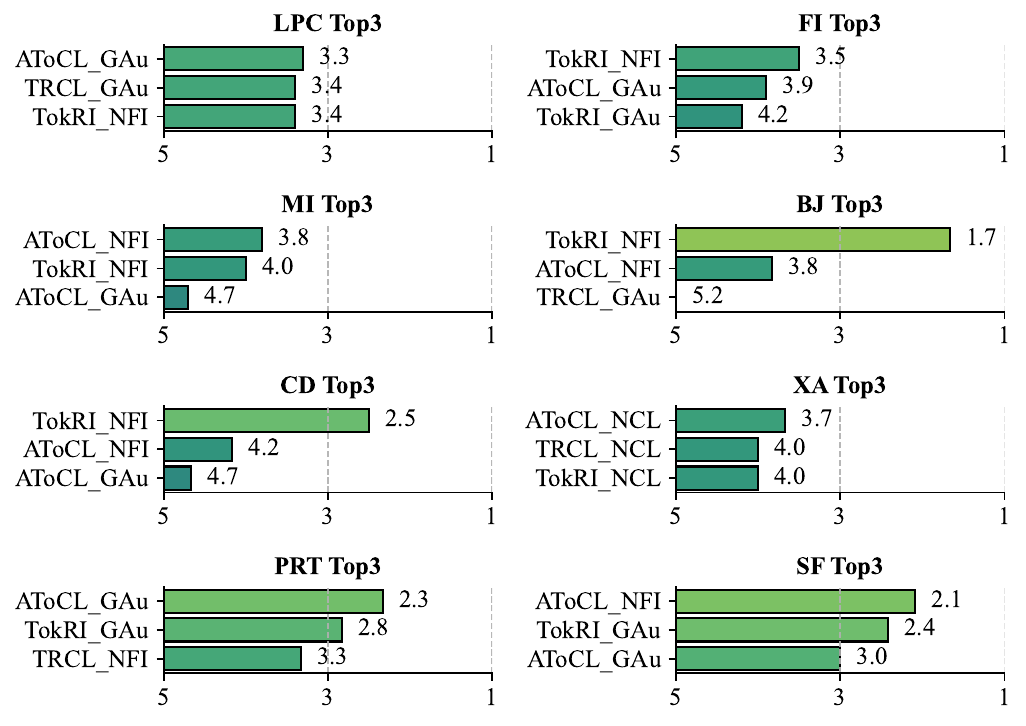}}
    \vspace{-2mm}
    \caption{Top 3 pretraining task combinations.}
    \label{fig:pretraintask}
    \vspace{-3mm}
\end{figure*}

\updatetw{
\section{Delving into Pretraining Tasks.}\label{sec:pretraintask}
In our experimental analysis in Section.\ref{sec:experiment}, we conclude that utilizing multiple pretraining tasks can enhance the performance of MapRL models. Moreover, greater task heterogeneity—incorporating token-based, graph-based, and sequence-based pretraining tasks—can further improve model capability. To fully investigate the influence of task heterogeneity on performance, we conduct deep dive experiments to figure out what specific combinations of heterogeneous pretraining tasks can yield the best gains on MapRL.

\paratitle{Experimental Space.} In this section, we aim to identify the best pretraining task combinations for MapRL. The candidate combination set is constructed by combining different categories of pretraining tasks. For POI entities, we consider token-based pretraining tasks (TokRI, TRCL) and sequence-based pretraining tasks (TrajP, MTR, ATrCL), forming combinations such as TokRI+TrajP, TRCL+MTR, \etc. Similarly, for road segments and land parcels, candidate sets are formed by combining token-, graph-, and sequence-based pretraining tasks (if applicable), as shown in Table~\ref{tab:pretraining_tasks}.

\begin{table}[t]
    \caption{Available pretraining tasks by entity types.}
    \centering
    \resizebox{0.9\linewidth}{!}{
        \setlength{\tabcolsep}{8pt}       
        \begin{tabular}{r|ccc}
        \toprule
        Task Type & POI   & Road Segment & Land Parcel \\
        \midrule
        Token-based & TokRI, TRCL & TokRI, TRCL & TokRI, TRCL, AToCL \\
        Graph-based & --     & NFI, GAu,AGCL   & NFI, GAu,NCL \\
        Seq-based & TrajP, MTR, ATrCL & TrajP, MTR, ATrCL & -- \\
        \bottomrule
        \end{tabular}%
    }
    \vspace{-6mm}
    \label{tab:pretraining_tasks}
\end{table}

The candidate set is constructed based on three rules: (1) selecting pretraining tasks implemented and validated by existing models; (2) using available datasets for testing; and (3) repeating all experiments five times with different random seeds. To ensure experimental feasibility, we further narrow the experimental scope by selecting only one pretraining task from each category for candidate combinations. The candidate set for other entity types follows the same principle.


\paratitle{Model Architecture.} To eliminate the influence of model architecture, we use the same encoder within each type of MapRL model and include only the necessary encoding components. Specifically, token-based encoders map entity IDs to embeddings. Additionally, the token encoder for POI incorporates spatial features by transforming coordinates into a vector using an MLP, while the token encoder for land parcel integrates semantic features in a similar way. Graph-based encoders use GCNs to process relation networks. For land parcels, the relation network defines edges using POI similarity, spatial distance, and human mobility transitions, following HREP~\cite{HREP}. Sequence-based encoders leverage Transformer layers to encode sequential auxiliary data. The model is configured with the same setting as the main experiment in Section~\ref{sec:experiment}. Encoders are arranged in the order: Token → Graph → Sequence.

\paratitle{Experimental Result.} 
To ensure a focused discussion, we highlight the top three pretraining task combinations for each entity type, ranked by their average performance across tasks and datasets. The results are shown in \fig~\ref{fig:pretraintask}, where each subfigure corresponds to a specific downstream task or dataset. The horizontal axis represents rankings (lower is better), while the vertical axis lists the top three combinations, with lighter colors indicating lower ranks. In \fig~\ref{fig:pretraintask}(a) for POI entities, the first three subfigures show each model’s average ranking in a specific downstream task (POIC, NPP, or TUL) across multiple datasets. We refer to these as {\em task-oriented rankings}. In contrast, the last five subfigures show each model’s average ranking across all tasks within a specific dataset (NY, TYO, CHI, SIN, or SF), which we refer to as {\em dataset-oriented rankings}. The same ranking logic applies to \fig~\ref{fig:pretraintask}(b) for road segments and \fig~\ref{fig:pretraintask}(c) for parcels. 
Based on the results, we have several key findings:

1) {\em Optimal Combinations for POI:}
TokRI + MTR consistently ranks highest, likely due to its ability to model POI categories and long-range temporal dependencies—two key attributes for POIs. However, in SF, where data sparsity is more pronounced, TokRI + TrajP emerges as a more suitable alternative, as TrajP focuses on local temporal patterns, making it more reliable in sparse data scenarios.

2) {\em Optimal Combinations for Road Segment:}
AGCL outperforms others in all scenarios, as it effectively captures important spatial dependencies in geographic relation (GR) networks, which is critical for modeling road segments. From a dataset-oriented perspective, AGCL + MTR achieves the best results in CD and PRT, where visit frequencies are higher (CD: 1551, PRT: 2430), allowing MTR to learn sequential dependencies from dense trajectory data. In contrast, AGCL + TrajP works better in BJ and SF, where visit frequencies are lower (BJ: 856, SF: 771). This suggests that MTR benefits more from dense data, while TrajP is better suited for sparse data.

3) {\em Optimal Combinations for Land Parcel:}
AToCL achieves the best performance in 5/8 cases, likely because its augmentation technique enhances the model’s ability to capture key semantic features while filtering out noise. This is especially important for land parcels, where POI category distributions define functional attributes. GAu and NFI are also well-suited for land parcels, as they better model similarity-driven connectivity in social relation (SR) networks. 

By integrating these insights, we provide a general guide for selecting pretraining tasks across different entity types:

1) {\em Token-based}: Encoding semantic features is a key capability of token-based encoders. TokRI is well-suited for structured categorical attributes, while AToCL’s augmentation-based contrastive learning better handles high-dimensional and redundant semantic features, making it particularly effective for land parcels.

2) {\em Graph-based}: AGCL is well-suited for geographic relation networks, as it effectively captures key spatial dependencies. In contrast, GAu and NFI perform better in social relation networks due to their ability to model similarity-driven connectivity.

3) {\em Sequence-based}: MTR excels at capturing long-range sequential dependencies, making it the best choice for sequence-based tasks, but it requires dense data to be effective. In contrast, TrajP is more reliable in sparse data scenarios.
}

\section{related work}\label{sec:relatedwork}
\updateo{
\paratitle{Niche Tasks for \rl.} Raster and vector data models are the two primary formats for electronic maps. The raster model structures spatial information through grids, which are commonly used in remote sensing and environmental analysis. In contrast, the vector model represents map entities as points, polylines, and polygons, leveraging graph structures to represent relations between entities. These structural differences lead to different modeling approaches: CNNs are usually used for raster-based learning, while GNNs better capture spatial relations in vector data. As a result, raster and vector models have distinct benchmarks. Due to its high precision and accuracy, the vector data model serves as the fundamental representation for electronic maps.} 

\updatetw{
In parallel, various downstream tasks in MapRL have been widely explored, including Next POI Prediction (NPP), Travel Time Estimation (TTE), and Mobility Inference (MI). These tasks play a key role in location-based services and spatiotemporal data mining. For NPP, models like DeepMove\cite{deepmove} employ hierarchical RNNs with attention mechanisms to capture sequential patterns in user trajectories. TTE methods such as DeepTTE \cite{deeptte} integrate LSTM networks with dynamic graph convolutional layers, fusing traffic state and weather data for accurate time estimation. For MI, models like ODMP~\cite{ODMP} adopt a full-pipeline approach, integrating sequence and graph encoders to effectively capture demand-supply dynamics in ride-hailing scenarios.
}

\paratitle{Reviews for \rl.} Pervious reviews~\cite{ST1, ST2, ST3} discuss the applications of deep learning in urban data mining, providing a detailed overview of data, tasks, and deep learning models, particularly those that integrate sequence and graph encoders. However, their focus is on end-to-end models, while our work concentrates on pre-trained representation learning methods. Other studies~\cite{tj1, tj2} focus on data mining methods for trajectory data but do not cover pre-trained representation learning. Survey~\cite{chen2024self} introduces a type-based taxonomy for \rl where models are systematically reviewed based on the types of map entities they target. In contrast, our work does not classify methods by entity type but instead focuses on summarizing general techniques applicable to all map entities, \ie a method-based taxonomy. This broader perspective helps uncover common design principles underlying \rl methods.

\paratitle{Benchmark for \rl.}
The growing interest in \rl has increased the demand for open benchmarks to analyze baseline models. Open-source benchmarking is valuable not only for advancing research but also for enabling users to evaluate models and apply them to open datasets. While significant progress has been made in benchmarking for multivariate time series prediction~\cite{libcity, TFB, tslib}, computer vision~\cite{imagenet}, and neural language processing~\cite{squad}, benchmarking in the \rl field remains largely unexplored. To our knowledge, \name is the first \rl benchmark for comparative experiments and model development.

\section{Conclusion}\label{sec:conclusion}

To overcome the challenges of fragmentation and the lack of standardized benchmarks in \rl, we introduced a novel taxonomy organized by functional modules rather than entity types. Building on this taxonomy, we developed \name, a modular library offering interfaces for encoding, pre-training, fine-tuning, and evaluation. Using \name, we established the first standardized benchmarks by reproducing 21 mainstream models and integrating datasets from nine cities. Moreover, we also conduct an in-depth evaluation of mainstream MapRL models. Our comprehensive experiments highlight the impact of pre-training tasks and encoder architectures, demonstrating the advantages of combining multiple components. VecCity provides a unified framework that promotes reusability, streamlines experimentation, and advances research in MapRL. Although \name currently supports most mainstream \rl models, we plan to expand it further to more pre-trained spatiotemporal data representation learning algorithms.


\clearpage

\bibliographystyle{ACM-Reference-Format}
\bibliography{sample}

\end{document}